\documentclass[nohyperref]{article}

\usepackage[utf8]{inputenc} 
\usepackage[T1]{fontenc}    
\usepackage{hyperref}       
\usepackage{url}            
\usepackage{booktabs}       
\usepackage{amsfonts}       
\usepackage{nicefrac}       
\usepackage{microtype}      
\usepackage{xcolor}         
\usepackage{amsmath}
\usepackage{amssymb}
\usepackage{amsthm}
\usepackage{mathtools}
\usepackage{bbm}
\usepackage{graphicx}
\usepackage[makeroom]{cancel}
\usepackage{subfig}
\usepackage{nicefrac}
\usepackage{caption}
\usepackage{mwe}
\usepackage{sidecap}
\usepackage{microtype}
\usepackage{multicol}

\usepackage[most]{tcolorbox}

\usepackage{hyperref}
\usepackage{scalerel}
\usepackage{stackengine,wasysym}

\usepackage[symbol]{footmisc}

\usepackage{amssymb}
\usepackage{pifont}
\newcommand{\cmark}{\ding{51}}%
\newcommand{\xmark}{\ding{55}}%

\usepackage[shortlabels]{enumitem}

\theoremstyle{plain}

\theoremstyle{definition}

\theoremstyle{remark}

\definecolor{lightgray}{HTML}{EEEEEE}
\definecolor{darkgray}{HTML}{666666}

\usepackage{amsmath,amsfonts,bm}

\def\eqref#1{equation~\ref{#1}}

\def\1{\bm{1}}

\def\rq{{\textnormal{q}}}

\def\rva{{\mathbf{a}}}

\def\rvd{{\mathbf{d}}}

\def\rvh{{\mathbf{h}}}

\def\rvp{{\mathbf{p}}}
\def\rvq{{\mathbf{q}}}

\def\rvt{{\mathbf{t}}}

\def\rvx{{\mathbf{x}}}

\def\rvz{{\mathbf{z}}}

\def\rmA{{\mathbf{A}}}

\def\rmD{{\mathbf{D}}}

\def\rmQ{{\mathbf{Q}}}

\def\rmZ{{\mathbf{Z}}}

\def\vq{{\bm{q}}}

\DeclareMathAlphabet{\mathsfit}{\encodingdefault}{\sfdefault}{m}{sl}
\SetMathAlphabet{\mathsfit}{bold}{\encodingdefault}{\sfdefault}{bx}{n}

\def\gT{{\mathcal{T}}}

\def\gX{{\mathcal{X}}}

\def\sA{{\mathbb{A}}}

\def\sD{{\mathbb{D}}}

\def\sR{{\mathbb{R}}}
\def\sS{{\mathbb{S}}}

\def\sX{{\mathbb{X}}}

\def\sZ{{\mathbb{Z}}}

\newcommand{\KL}{D_{\mathrm{KL}}}

\newcommand{\defeq}{\vcentcolon=}

\newcommand{\Ex}{\mathbb{E}}

\newcommand{\iidsim}{\overset{\mathrm{iid}}{\sim}}
\newcommand{\prioritysim}{\overset{\mathrm{priority}}{\sim}}
\newcommand{\RVB}{\mathcal{L}}
\newcommand{\VIB}{\mathcal{L}_\mathrm{ELBO}}

\newcommand{\ind}{\mathbbm{1}}

\newcommand{\pth}{p_\theta}

\newcommand{\rphi}{r_\phi}

\newcommand{\wall}{w_{\theta,\phi}}
\newcommand{\vall}{v_{\theta,\phi}}

\newcommand{\fth}{f_\theta}
\newcommand{\gth}{g_\theta}

\newcommand{\fphi}{f_\phi}

\newcommand{\bert}{\mathrm{BERT}}

\newcommand{\linear}{\mathrm{Linear}}
\newcommand{\EMDR}{\textsc{Emdr-2}}

\usepackage{hyperref}

\usepackage[accepted]{icml2023}

\usepackage[capitalize,noabbrev]{cleveref}

\usepackage[textsize=tiny]{todonotes}

\newcommand{\refp}[1]{(\ref{#1})}

\icmltitlerunning{Variational Open-Domain Question Answering}

\begin{document}

\twocolumn[
\icmltitle{Variational Open-Domain Question Answering}



\icmlsetsymbol{equal}{*}

\begin{icmlauthorlist}
\icmlauthor{Valentin Liévin}{1,2}
\icmlauthor{Andreas Geert Motzfeldt}{1}
\icmlauthor{Ida Riis Jensen}{1}
\icmlauthor{Ole Winther}{1,2,3,4}
\end{icmlauthorlist}

\icmlaffiliation{1}{Section for Cognitive Systems, Technical University of Denmark, Denmark}
\icmlaffiliation{2}{FindZebra, Denmark}
\icmlaffiliation{3}{Center for Genomic Medicine, Rigshospitalet, Copenhagen University Hospital, Denmark}
\icmlaffiliation{4}{Bioinformatics Centre, Department of Biology, University of Copenhagen, Denmark}
\icmlaffiliation{4}{Bioinformatics Centre, Department of Biology, University of Copenhagen, Denmark}

\icmlcorrespondingauthor{Valentin Liévin}{valv@dtu.dk}

\icmlkeywords{Machine Learning, Variational Inference, Optimization, Discrete, Latent Variable, Question Answering, Information Retrieval, Natural Language Processing, Language Modelling, Medical, MedMCQA, USMLE}

\vskip 0.3in
]



\printAffiliationsAndNotice{\icmlEqualContribution} 

\begin{abstract}
Retrieval-augmented models have proven to be effective in natural language processing tasks, yet there remains a lack of research on their optimization using variational inference. We introduce the Variational Open-Domain (VOD) framework for end-to-end training and evaluation of retrieval-augmented models, focusing on open-domain question answering and language modelling. The VOD objective, a self-normalized estimate of the Rényi variational bound, approximates the task marginal likelihood and is evaluated under samples drawn from an auxiliary sampling distribution (cached retriever and/or approximate posterior). It remains tractable, even for retriever distributions defined on large corpora. We demonstrate VOD's versatility by training reader-retriever BERT-sized models on multiple-choice medical exam questions. On the MedMCQA dataset, we outperform the domain-tuned Med-PaLM by +5.3\% despite using 2.500$\times$ fewer parameters. Our retrieval-augmented BioLinkBERT model scored 62.9\% on the MedMCQA and 55.0\% on the MedQA-USMLE. Last, we show the effectiveness of our learned retriever component in the context of medical semantic search.
\end{abstract}

\section{Introduction}
\label{submission}

\begin{figure}[ht]
\begin{center}
\centerline{\includegraphics[width=1.0\columnwidth]{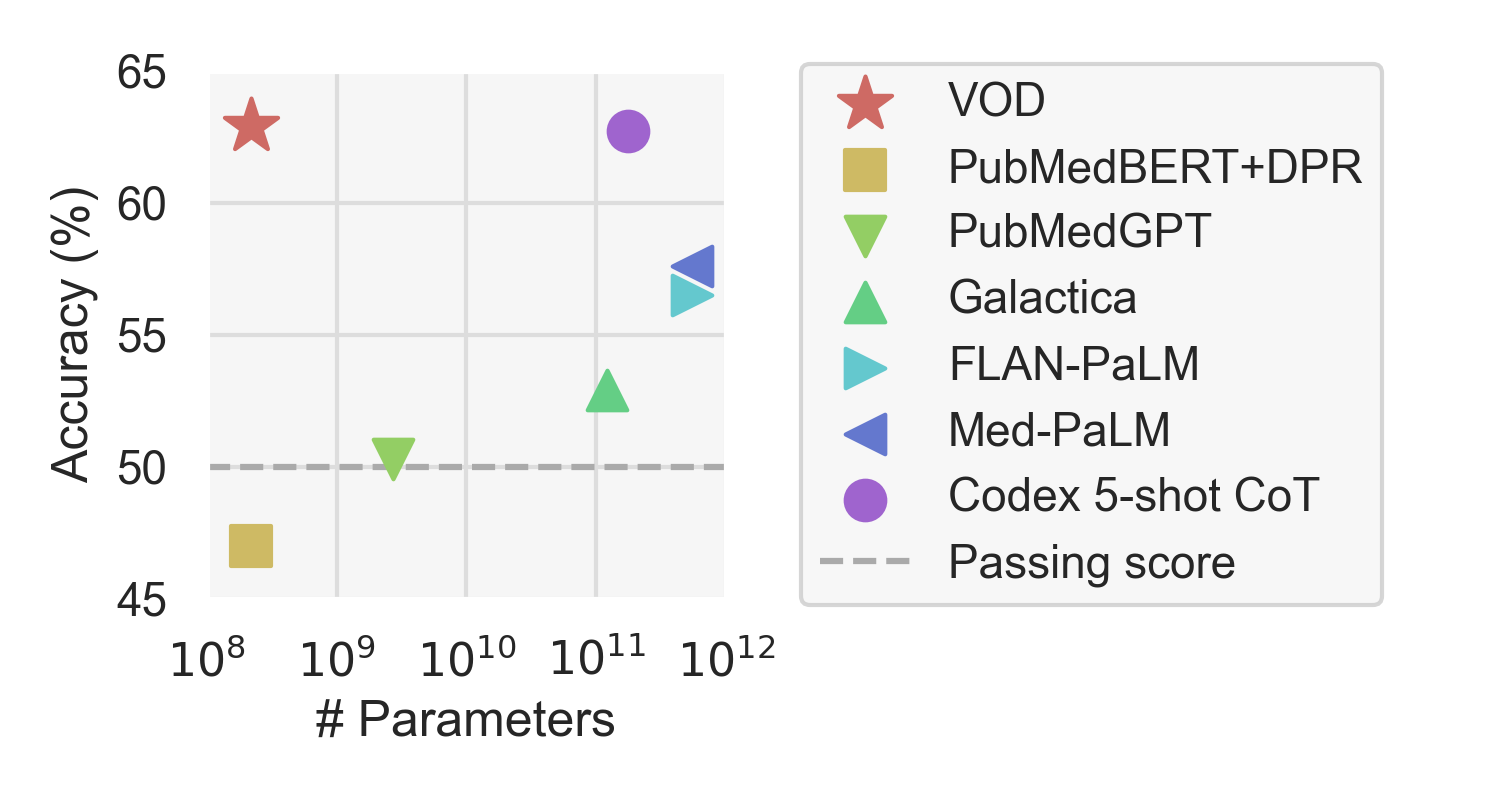}}
\caption{Parameter efficiency. Answering accuracy of baseline methods and of VOD (BioLinkBERT backbone) on MedMCQA.}
\label{fig:vod-benchmark}
\end{center}
\vskip -0.2in
\end{figure}


Scaling Transformer-based \citep{Vaswani2017-st} language models (LMs) with larger datasets and more parameters \citep{Radford2018-kq, Kaplan2020-hw, Hoffmann2022-qq} led to sustained improvements in various downstream tasks.\footnote{\scriptsize Find a benchmark of LLMs in \citet{Srivastava2022-jl}, read about LLMs in \citet{Brown2020-ad, Rae2021-oy, Chowdhery2022-pr, Thoppilan2022-tm, Hoffmann2022-qq, Smith2022-jc, Zhang2022-zr, Lieber2021-gw, Fedus2021-ci, Laurencon2022-rx}.} 
However, large language models (LLMs) may reach a plateau in their performance due to the limitations of the implicit knowledge they possess, being incomplete, flawed or out-of-date. \textit{Open-domain question answering} (ODQA) consists of augmenting LMs with external knowledge bases indexed with a retrieval mechanism. This approach was popularized in the question-answering setting by \citet{Chen2017-av} and was later applied to the task of language modelling itself \citep{Guu2020-la, Lewis2020-cg, Borgeaud2021-td, Izacard2022-co}.

However, optimizing deep retrievers is challenging, unless there is a set of annotated evidence documents that are sufficiently aligned with the target task, as explored in \citet{Karpukhin2020-di, Qu2021-aq, Khattab2020-tn}. An alternative approach is to model the whole collection of documents as a latent variable \cite{Lee2019-mr}, but this still poses challenges for optimization, especially considering that documents are discrete quantities.\footnote{\scriptsize Learn more about discrete latent variable optimization in \citet{Hinton1995-qk, Le2018-kj, Mnih2014-ev, Mnih2016-vr, Van_den_Oord2017-sg, Tucker2017-vy, Grathwohl2017-hm, Masrani2019-nq, Lievin2020-cm}.}

This research fills a gap in the literature by exploring the optimization of retrieval-augmented models using variational inference. We introduce a probabilistic framework that extends Rényi divergence variational inference~\citep{Li2016-ph}, allowing us to estimate the marginal task likelihood and its gradient by sampling from an \textit{approximate posterior}. The proposed framework is versatile and applies to various settings, including extractive, generative, and multiple-choice models for open-domain question answering, as well as the training of retrieval-enhanced language models. 

To demonstrate the effectiveness of the framework, we train reader-retriever BioLinkBERT models end-to-end on multiple-choice medical QA tasks and achieve a new state-of-the-art on the MedMCQA of $62.9$\%, outperforming the current 540B parameter domain-tuned Med-PaLM by +5.3\% \cite{Singhal2022-la} using 2.500$\times$ fewer parameters (Figure \ref{fig:vod-benchmark}). On the challenging MedQA-USMLE, we score $55.0$\%: a new state-of-the-art in the open-domain setting. We highlight the main contributions of this paper as follows:

\begin{enumerate}[itemsep=0pt,topsep=0pt]
\item The VOD framework: tractable, consistent, end-to-end training of retrieval-augmented models.
\item Popularizing Rényi divergence variational inference for natural language tasks.
\item Truncated retriever parameterization: relaxing the top-$K$ retriever approximation to using top $P \geq K$.
\end{enumerate}
In addition to our theoretical contributions,  we release MedWiki: a subset of Wikipedia tailored to the MedMCQA and USMLE dataset for low-resource research.



\section{VOD: a Probabilistic Framework for Retrieval-augmented Tasks}

Let a question $\rvq$ be defined in a space $\Omega$ (e.g., the space of sequences of tokens) and the set of possible answers be $\sA \subset \Omega$ with a correct answer denoted $\rva \in \sA$. We introduce a corpus of $N$ documents $\sD \defeq \{\rvd_1,\ldots,\rvd_N \} \in \Omega^N$. In open-domain tasks, we are interested in modelling the marginal task likelihood with a reader-retriever model $\pth(\rva, \rvd | \rvq) \defeq \pth(\rva | \rvd, \rvq)\pth(\rvd | \rvq)$ parameterized by $\theta$:
\begin{equation}\label{eq:marginal-task-likelihood}
    \pth(\rva | \rvq) \defeq \sum_{\rvd \in \sD}  
    \underbrace{\pth(\rva | \rvd, \rvq)}_{\text{reader}}
    \,
    \underbrace{\pth(\rvd | \rvq)}_{\text{retriever}}
    \,.
\end{equation}
%


Variational inference \cite{Jordan1999-qb, Kingma2013-ey, Burda2015-wt} allows estimating the marginal task likelihood eq. \refp{eq:marginal-task-likelihood} using samples drawn from an approximate posterior $\rphi(\rvd | \rva, \rvq)$. This consists of evaluating the \textit{evidence lower bound} (ELBO), a log-likelihood lower bound.\footnote{{\scriptsize $\VIB(\rva, \rvq) \defeq \log \pth(\rva, \rvq) - \mathcal{D}_\mathrm{KL}\left( \rphi(\rvd | \rva, \rvq) \| \pth(\rvd | \rva, \rvq)\right)$}} In open-domain applications, the approximate posterior, with parameter $\phi$, can be defined using either a keyword-search engine (BM25; \citet{Robertson2009-gm}), a checkpoint of $\pth(\rvd | \rvq)$, or a model learned jointly. 

We introduce the VOD framework in four acts: i) Why Rényi divergence variational inference can aid likelihood-based learning, ii) The VOD objective: a tractable self-normalized importance sampling estimate of the Rényi bound, iii) A truncated retriever parameterization that generalizes existing approaches and iv) A discussion on the application of the VOD framework.



\subsection{Rényi Divergence Variational Inference}

\begin{figure}[ht]
\vskip 0.25in
\begin{center}
\centerline{\includegraphics[width=1.0\columnwidth]{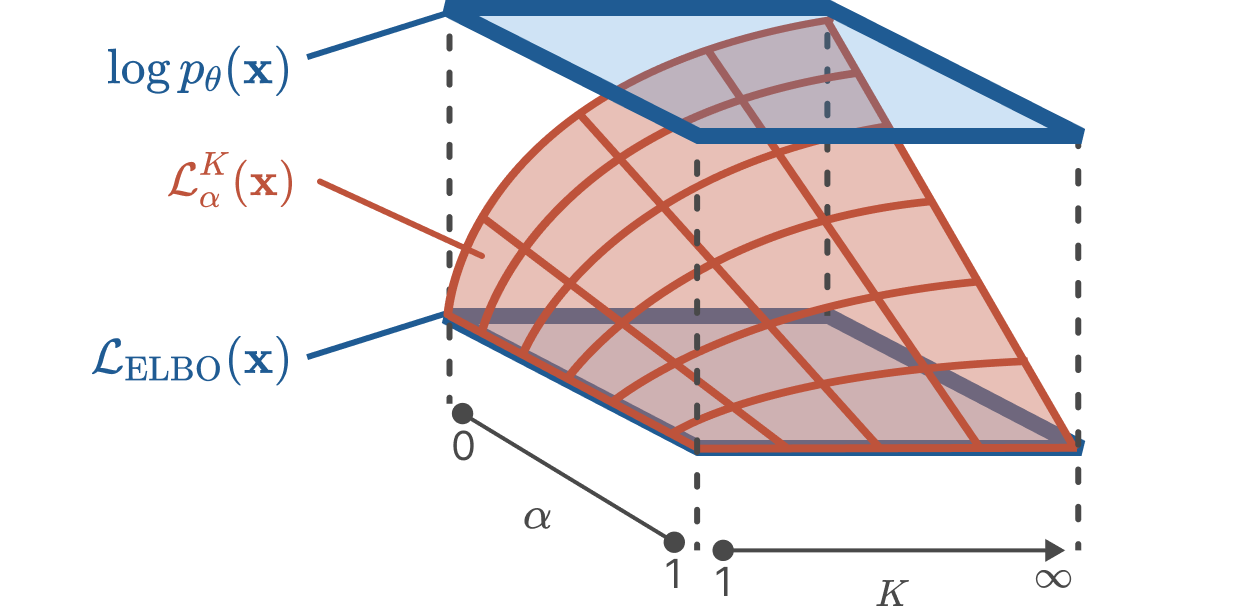}}
\caption{Depicts the core component of the VOD framework: the importance-weighted Rényi Variational Bound (IW-RVB) as a function of the parameter $\alpha \in [0,1]$ and the number of samples $K \geq 1$. As the value of $\alpha$ and $K$ increase, the IW-RVB becomes a more accurate estimate of the likelihood of a given task, demonstrating how we use VOD to optimize retrieval-augmented models through the manipulation of $\alpha$ and $K$. See how the parameter $\alpha$ affects the training dynamics in Figure \ref{fig:apdx-experimental-kl}, Appendix \ref{apdx:experimental-results}.}

\label{fig:rvb-3d}
\end{center}
\vskip -0.05in
\end{figure}

Rényi divergence variational inference \citep{Li2016-ph} extends traditional variational inference \cite{Jordan1999-qb, Kingma2013-ey}. Given a parameter $\alpha < 1$ and the importance weight $\wall^{1-\alpha}(\rva, \rvq, \rvd) \defeq \pth(\rva, \rvd | \rvq)\rphi^{-1}(\rvd | \rva, \rvq)$ the variational R\'enyi bound (RVB) defined as
\begin{equation}\label{eq:renyi-bound}
    \RVB_\alpha(\rva, \rvq) \defeq \frac{1}{1-\alpha} \log \Ex_{\rphi(\rvd | \rva, \rvq)} \left[ \wall^{1-\alpha}(\rva, \rvq, \rvd)\right] 
\end{equation}
RVB is a lower bound of the marginal log-likelihood for $\alpha \geq 0$ and is extended by continuity in $\alpha=1$ as $\RVB_{\alpha = 1}(\rva, \rvq) \defeq \lim_{\alpha \rightarrow 1}\RVB_{\alpha}(\rva, \rvq)$ where it equals the ELBO.
In practice, the RVB and its gradients can be estimated using $K$ documents sampled from $\rphi(\rvd|\rva, \rvq)$. The resulting importance sampling estimate yields another bound: the Importance Weighted RVB (IW-RVB; \citet{Li2016-ph}):
%
\begin{gather}\label{eq:iw-rvb}
    \hat{\mathcal{L}}_\alpha^K(\rva, \rvq) \defeq
    \frac{1}{1-\alpha} \log \frac{1}{K} \sum_{i=1}^K \wall^{1-\alpha}(\rva, \rvq, \rvd_i) \\
    \rvd_1, \ldots, \rvd_K \iidsim \rphi(\rvd | \rva, \rvq) \nonumber
\end{gather}
%
which aligns with the importance-weighted bound (IWB; \citet{Burda2015-wt}) in $\alpha=0$. To sum up, the main properties of the RVB and the IW-RVB are ($\alpha \geq 0$):
\begin{subequations}
\begin{align*}
   \RVB_{\alpha = 0}(\rva, \rvq) =& \log \pth(\rva | \rvq) \,&
  \RVB_{\alpha \rightarrow 1}(\rva, \rvq)  = & \VIB(\rva, \rvq) \\
  \RVB_{\alpha \geq 0}(\rva, \rvq) \leq & \log \pth(\rva | \rvq) \,&
  \RVB_{\alpha}^K(\rva, \rvq)  \leq & \RVB_{\alpha}(\rva, \rvq)  \ .
\end{align*}
\end{subequations}


\paragraph{RVB gradient}

The gradient of the RVB w.r.t. $\theta$ is:
\begin{equation*}\label{eq:renyi-gradient}
\resizebox{\columnwidth}{!}{
    $\nabla_\theta \RVB_{\alpha}(\rva, \rvq) = \Ex_{\rphi} \left[\widetilde{\wall^{1-\alpha}}(\rva, \rvq, \rvd) \ \nabla_\theta\log \pth(\rva, \rvd | \rvq) \right]$
    }
\end{equation*}
where the normalized importance weight is defined as 
\begin{equation}
\widetilde{\wall^{1-\alpha}}(\rva, \rvd) \defeq \frac{\wall^{1-\alpha}(\rva, \rvq, \rvd) }{\Ex_{\rphi(\rvd' | \rva, \rvq)} \left[ \wall^{1-\alpha}(\rva, \rvd', \rvq) \right]}\,.
\end{equation}

In this paper, we consider the sampling distribution $\rphi$ to be static and therefore do not estimate the gradient w.r.t. the approximate posterior. Optimizing the parameter $\phi$ jointly with $\theta$ can be done by application of importance sampling coupled with variance reduction techniques \citep{Burda2015-wt, Mnih2016-vr, Le2018-kj, Masrani2019-nq, Kool2019-qw, Lievin2020-cm}.

\paragraph{Stabilizing training using the RVB}

Considering the optimization of the parameter $\phi$, a looser bound (e.g., the ELBO) might be preferred to a tighter one (e.g., the IWB).\footnote{Exploring using hybrid ELBO/IWB objectives has been explored in \citet{Rainforth2018-pw}, interpolating the RVB has been explored in \citet{Lievin2020-cm}.} In this paper, we explore interpolating between variational bounds using the parameter $\alpha$ of the RVB. We argue that, even for a non-trainable parameter $\phi$, optimizing for a looser bound can overcome early optimization challenges.

For $\alpha = 0$, the RVB aligns with the marginal log-likelihood independently of the choice of the approximate posterior. However, when the importance weight $\wall(\rvq, \rva, \rvd)$ suffers from high variance, so does the Monte Carlo estimate of the marginal likelihood and its gradient.\footnote{See \citet{Kong1992-mz, Owen2013-hr, Nowozin2015-eo} for an introduction and discussion about variance and importance sampling.}


For $\alpha=1$, the RVB matches the ELBO and the gradients restricted to the reader and retriever decomposes as:
\begin{align}
\nabla_{\theta_{\textsc{\scriptsize Read.}}} \RVB_{\alpha=1}(\rva, \rvq) & = \Ex_{\rphi(\rvd | \rva, \rvq)} \left[ 
   \nabla_\theta \log \pth(\rva | \rvd, \rvq) 
  \right] \nonumber\\
\nabla_{\theta_{\textsc{\scriptsize Retr.}}} \RVB_{\alpha=1}(\rva, \rvq) & = - \nabla_\theta \KL\left( \rphi(\rvd | \rva, \rvq) \|  \pth(\rvd | \rvq) \right) \nonumber  \ .
\end{align}
Maximizing the ELBO corresponds to optimizing the reader and the retriever disjointly. On the reader side, this equals maximizing the answer likelihood $\pth(\rva | \rvd, \rvq)$ in expectation over $\rphi(\rvd|\rva, \rvq)$ independently of the value of $\pth(\rvd | \rvq)$. On the retriever side, this corresponds to matching the approximate posterior with the learned retriever $\pth(\rvd | \rvq)$. This can be seen as an instance of knowledge distillation of the posterior into the retriever. After an initial learning phase, the RVB can be smoothly interpolated from the ELBO to the marginal task likelihood by controlling the parameter $\alpha$.

\begin{tcolorbox}[breakable,colback=lightgray,colframe=darkgray,left=4pt,right=4pt,top=4pt,bottom=4pt]
\subsection{VOD objective}

\vspace{0.8em}
In ODQA applications, the IW-RVB eq. \refp{eq:iw-rvb} is generally intractable due to the normalization constant in eq. \refp{eq:retriever-param} which requires evaluating all documents. 

\vspace{0.8em}
The VOD objective is an approximation of the IW-RVB which can be evaluated using $K$ documents sampled \textit{without replacement} from $\rphi(\rvd | \rva, \rvq)$. It is defined as:
%
\begin{gather}\label{eq:vod-objective}
\hat{L}^{K}_{\alpha}(\rva, \rvq) \defeq
    \frac{1}{1-\alpha} \log \sum_{i=1}^K s_i\, \hat{v}_{\theta,\phi}^{1-\alpha} (\rva, \rvq, \rvd_i) \\
    (\rvd_1, s_i), \ldots, (\rvd_K, s_K) \prioritysim \rphi(\rvd | \rva, \rvq) \nonumber \,.
\end{gather}%
%
where the self-normalized importance weight $\hat{v}_{\theta,\phi}$ is defined using the un-normalized retrieval density ratio $\zeta(\rvd) \propto \pth(\rvd|\rvq) \rphi^{-1}(\rvd| \rva, \rvq)$ as:
\begin{equation}\label{eq:vod-weight}\\ 
\hat{v}_{\theta,\phi} \defeq\, \pth(\rva| \rvq, \rvd_i) \zeta (\rvd_i) \left(\sum_{j =1}^K s_j \zeta(\rvd_j)\right)^{-1} \\
\end{equation}
The set of documents $\rvd_1, \ldots, \rvd_K$ are sampled without replacement from $\rphi(\rvd | \rva, \rvq)$ using \textit{priority sampling}~\citep{Duffield2007-rd}. The sampling procedure comes with importance weights $s_1, \ldots, s_k$ defined such that for a function $h(\rvd)$,  $\sum_{i=1}^K s_i\, h(\rvd_i) \approx \Ex_{\rphi(\rvd | \rva, \rvq)}\left[ h(\rvd) \right]$. We present priority sampling in greater length in Appendix \ref{apdx:priority-sampling}.

\vspace{0.8em}
The VOD objective and its gradient are consistent (i.e., converge to the RVB in the limit $K \rightarrow N$ with probability one) and can be evaluated with complexity $\mathcal{O}(K)$, whereas the IW-RVB is of complexity $\mathcal{O}(N)$. Furthermore, the VOD objective approximates the IW-RVB, which itself is guaranteed to approximate the marginal task log-likelihood more tightly as $K \rightarrow N$~\citep{Burda2015-wt}. 

\vspace{0.8em}
The VOD objective is derived in Appendix \ref{apdx:vod-objective}, the VOD gradient is defined in Appendix \ref{apdx:vod-gradients}. Our implementation of the sampling methods and the VOD objective is available at \url{http://github.com/VodLM/vod}.
\end{tcolorbox}

\subsection{Truncated retriever parameterization}

The VOD framework is compatible with retrievers defined on the whole corpus ($N$ documents). However, in our approach, we truncate the retriever to consider only the top $P$ documents, where $K < P \ll N$. $K$ refers to the number of sampled documents, while $P$ represents the pool of documents from which the top $K$ documents are selected. This truncation provides two key advantages: i) it enables efficient caching or retention of document scores, as only $P$ documents need to be stored in memory, and ii) the value $P$ serves as an exploration-exploitation threshold: a higher value of $P$ yield greater diversity in document sampling, promoting \textit{exploration}. While, a smaller value of $P$ ensures that during training, all documents in the set $\gT_\phi$ are more likely visited, facilitating \textit{exploitation} of the available information.

Assuming the retrieval distributions to be described by score functions $\fth: \Omega^2 \rightarrow \sR$ and $\fphi: \Omega^3 \rightarrow \sR$. We define the truncated retrievers as:\footnote{When $P > K$, evaluating the retriever density eq. \refp{eq:retriever-param} is generally intractable due to the sum over $P$ documents.}
\begin{subequations}\begin{align}
    \pth(\rvd | \rvq) \defeq& \frac{ \ind[\rvd \in \gT_\phi] \exp{\fth(\rvd, \rvq)}}{\sum_{\rvd' \in \gT_\phi} \exp{\fth(\rvd', \rvq)}} \label{eq:retriever-param} \\ 
    \rphi(\rvd | \rva, \rvq) \defeq& \frac{\ind[\rvd \in \gT_\phi] \exp{\fphi(\rva, \rvq, \rvd)}}{\sum_{\rvd' \in \gT_\phi} \exp{\fphi(\rva, \rvq, \rvd')}} 
\end{align}\end{subequations}
where $\gT_\phi$ is the set of the top $P \leq N$ documents ranked by the score $\fphi(\rva, \rvq, \rvd)$. The score function $\fth$ and $\fphi$ can be implemented using BM25 and/or contextual vector representations extracted using pretrained language models such as DPR or ColBERT \cite{Karpukhin2020-di, Khattab2020-tn}. For instance using a dual-encoder model $\fth(\rvd, \rvq) = \bert_\theta(\rvd)^T\bert_\theta(\rvq)$ and $\fphi(\rva, \rvq, \rvd) = \bert_\phi([\rvq ; \rva])^T \bert_\phi(\rvd)$ where $\bert$ is the function that return the output of a BERT model at the CLS token and $[\cdot ; \cdot]$ is the concatenation operator. Retrieving the top $P$ documents is efficient when using \texttt{elasticsearch}\footnote{\url{http://www.elastic.co/}} and/or \texttt{faiss}~\citep{Johnson2021-do}.

\subsection{Applying VOD}

In this paper, we show how to apply the VOD framework to multiple-choice ODQA. Nevertheless, VOD is general-purpose and designed for latent variable models defined on a discrete and finite space. In NLP, it applies to a wide range of settings such as generative, extractive, multiple-choice ODQA as well as retrieval-augmented language modelling. Find a non-exhaustive list of examples in Appendix \ref{apdx:applications}.

\section{Related work}


VOD aids the development of retrieval-augmented models for language modeling (LM) tasks. In this section, we review previous work on retrieval for LM, and compare to VOD (summarized with references in Table \ref{tab:related-work}).

\begin{table}[t]
\vspace{-2mm}
  \caption{Deep retrievers in literature, detailing if training was end-to-end, variational, as well the size of support during training.}
  \label{tab:related-work}
  \vskip 0.15in
  \begin{center}
  \resizebox{\columnwidth}{!}{\renewcommand{\arraystretch}{1.0} 
  \begin{tabular}{ll@{\hspace{-0.5em}}ccc}
    \toprule
    \bf Method & \bf Retriever training & \bf \shortstack{End-to-end\\learning} & \bf \shortstack{Posterior \\ Guided} & \bf \shortstack{Retriever \\ Support}    \\ 
    \midrule
    DPR\textsuperscript{1}& Supervised & \xmark & \xmark & --  \\
    ColBERT\textsuperscript{2} & Supervised & \xmark  & \xmark& -- \\
    Contriever\textsuperscript{3} & Self-supervised & \xmark & \xmark & --   \\
    FiD\textsuperscript{4} & Frozen DPR dual-encoder & \xmark & \xmark & --  \\
    RETRO\textsuperscript{5} & Frozen BERT dual-encoder & \xmark & \xmark & --  \\
    \addlinespace[0.2cm]
    ORQA\textsuperscript{6} & Self-supervised + MLL\textsuperscript{*} & (\cmark) & \xmark & top-$K$ doc. \\
    RAG\textsuperscript{7} & {MLL\textsuperscript{*}  + frozen DPR doc. encoder} & (\cmark) & \xmark  & top-$K$ doc. \\
    REALM\textsuperscript{8} & Self-supervised + MLL\textsuperscript{*} & \cmark & \xmark & top-$K$ doc.  \\
    $\EMDR$\textsuperscript{9} & {Self-supervised + Expect.-Max.}  & \cmark & \cmark & top-$K$ doc. \\
    Hindsight\textsuperscript{10}  & {ColBERT init. + ELBO + MLL\textsuperscript{*}} & \cmark & \cmark & top-$K$ doc.   \\
    \midrule
    VOD & Rényi variational bound & \cmark & \cmark & top-$P$ doc.\textsuperscript{$\dagger$}\\
    \bottomrule
    \addlinespace[0.1cm]
    \multicolumn{5}{l}{
    \small \textsuperscript{1}~\citet{Karpukhin2020-di},
    \textsuperscript{2}~\citet{Khattab2021-xu},
    \textsuperscript{3}~\citet{Izacard2021-te},
    \textsuperscript{4}~\citet{Izacard2020-ui}
    }\\
    \multicolumn{5}{l}{
    \textsuperscript{5}~\citet{Borgeaud2021-td},
    \textsuperscript{6}~\citet{Lee2019-mr},
    \textsuperscript{7}~\citet{Lewis2020-cg},
    \textsuperscript{8}~\citet{Guu2020-la}
    }\\
    \multicolumn{5}{l}{
    \textsuperscript{9}~\citet{Sachan2021-vq},
    \textsuperscript{10}~\citet{Paranjape2021-dp},
    \textsuperscript{*}MLL: marginal log-likelihood
    }\\
    \multicolumn{5}{l}{
    \small \textsuperscript{$\dagger$} $K \leq P \leq N$ ( $K$:\# of documents in a batch, $N$: corpus size, $P$: chosen)
    }
  \end{tabular}}
  \end{center}
  \vskip - 0.10in
\end{table}

\paragraph{Learning to search}

Retrieval-based training have gained much attention for improving pre-trained LMs. ORQA and Contriever proposed a self-supervised approach using contrastive learning to match a text passage with its context, and is widely adopted in pre-training to enable zero-shot retrieval (\textit{Inverse Cloze Task}; \citet{Lee2019-mr}). In contrast, DPR and ColBERT use supervised contrastive learning with questions paired to annotated documents. This method has sparked many retrieval-augmented attempts such as FiD, RETRO, and RAG to enhance auto-regressive LMs conditioned on a frozen retriever. ORQA and REALM, later followed by RAG, EMDR, Hindsight, and VOD proposed optimizing both a retrieval component and a reader or language modelling component end-to-end, by maximizing the marginal log-likelihood (MLL). 

\paragraph{Posterior guided supervision}

Many efforts has been devoted to leveraging external knowledge with posterior guided supervision. EMDR learns a retriever end-to-end with an Expectation-Maximization objective evaluated under the posterior distribution of $\pth(\rvd | \rva, \rvq) \propto \pth(\rvd | \rvq) \pth(\rva | \rvd, \rvq)$, while Hindsight optimizes the variational lower-bound (ELBO) evaluating under a target-aware approximate posterior $\rphi(\rvd | \rva, \rvq)$. Among previous methods, Hindsight is most akin to VOD as both methods rely on maximizing a variational bound. Nonetheless, VOD introduces the more general Rényi variational bound, which offers to model the sampling distribution explicitly. Ultimately, a more principled approach makes VOD more versatile and capable of handling a wider range of problems.



\paragraph{Navigating large knowledge bases}

The large size of knowledge bases such as Wikipedia makes it computationally intractable to consider all $N$ documents when computing MLL. To address this, all related methods rely on a strict truncation of the retriever to the top-$K$ cached documents. In contrast to these aforementioned approaches, which limits to a fixed set of $K$ documents, we propose a truncated retriever parameterization that works hand-in-hand with our principled objective to handle over top $P > K$ documents. Ultimately, this allows for more diverse document sampling during training and allows reducing the bias induced by truncating the retriever distribution. In Appendix \ref{apdx:realm}, we show that the top-$K$ MLL is a special case of VOD for $K=P$ and $\alpha=0$.




\section{Experiments}

In this section, we present the medical domain tasks and datasets, results on end-to-end multiple-choice ODQA and its application to information retrieval. The code and datasets are available on GitHub.\footnote{
\ifdefined\isaccepted
    \scriptsize \url{https://github.com/findzebra/fz-openqa}
\else
    \textit{GitHub repository to be disclosed upon publication.}
\fi
}

\subsection{Datasets}

\begin{table}[t]
\vspace{-2mm}
\caption{Summarizes the medical QA datasets and corpora used in our study, including the MedMCQA, USMLE, and FindZebra (FZ) corpus, with the MedWiki as the knowledge base for all QA tasks. The questions are numbered for the train/validation/test splits.}
\label{tab:datasets}
\begin{center}
\begin{small}
\begin{sc}
\vskip -0.1in
\resizebox{\columnwidth}{!}{
\begin{tabular}{lccc}
\toprule
\textbf{Datasets}  & \textbf{MedMCQA} & \textbf{USMLE} & \textbf{FZ queries} \\
\addlinespace[0.3em]
questions  & 182.8k/4.2k/6.1k  & 10.2k/1.3k/1.3k & 248   \\
\midrule
\textbf{Corpora} & \textbf{Wikpedia} & \textbf{MedWiki} & \textbf{FZ corpus} \\
\addlinespace[0.3em]
Articles  & 6.6M & 293.6k & 30.7k  \\
Passages & -- & 7.8M & 711.9k \\
\bottomrule
\end{tabular}
}
\end{sc}
\end{small}
\end{center}
\vskip -0.15in
\end{table}

The datasets utilized for the medical domain are summarized in Table \ref{tab:datasets}. We introduce the MedWiki, a subset of Wikipedia targeted to medical QA tasks. 

\paragraph*{MedMCQA}
\citet{Pal2022-ph} is a large-scale multiple-choice question answering dataset collected from Indian medical school entrance exams (AIIMS and NEET-PG). It covers several medical topics (dentistry, pathology, surgery, preventive medicine, etc.) and question types (diagnosis, recalling expert factual knowledge, mathematical problems, etc.)

\paragraph*{MedQA-USMLE}
\citet{Jin2021-jo}) is a collection of medical questions from the US medical board exam. The questions aim to assess human doctors' medical knowledge and decision-making. Each question includes a medical history, vital signs (e.g., blood pressure, temperature), and possibly a specific analysis (e.g., CT scan).

\paragraph*{MMLU}
\citet{hendrycks2021measuring} is a dataset for assessing the knowledge acquired during pre-training by evaluating models in a zero-shot setting. The test set comprises 57 tasks spanning different domains. We limit our analysis to the subcategories \textit{psychology}, \textit{biology}, and \textit{health}.\footnote{The subcategory \textit{professional\_medicine} corresponds to the MedQA-USMLE questions.}

\paragraph*{MedWiki}
We release the MedWiki corpus (under MIT license): a collection of $4.5\%$ of articles taken from the English Wikipedia and targeted to the MedMCQA and USMLE datasets. The MedWiki corpus was built by querying each answer option from the MedMCQA and USMLE datasets against the Wikipedia API. 
Read more in Appendix \ref{apdx:medwiki}. 

\paragraph*{FindZebra corpus \& queries}

FindZebra is a search tool for assisting in the diagnosis of rare diseases that is built on open-source information retrieval software (BM25) tailored to this problem \cite{Dragusin2013-jw}. The FindZebra corpus indexes a collection of curated articles from various reputable databases: GARD, GeneReviews, Genetics Home Reference, OMIM, Orphanet, and Wikipedia. Each article is referenced with a Concept Unique Identifier (CUI) from the Unified Medical Language System (UMLS; \citet{Bodenreider2004-vo}). We use a collection of 248 publicly available search queries (FZ queries). Each query is labelled with a reference diagnostic, allowing to benchmark medical search engines.\footnote{\scriptsize \url{https://huggingface.co/datasets/findzebra}}

\subsection{VOD for multiple-choice QA}

In the multiple-choice question answering (MCQA) setting, we consider a vector of $M$ answer options $\rmA = [\rva_1, \ldots, \rva_M]$, where $\star$ represents the index of the correct option. Similarly, we define a vector of $M$ queries as $\rmQ = [\rvq_1, \ldots, \rvq_M]$, where $\rvq_j = [\rvq; \rva_j]$ represents the concatenation of the question with the answer option of index $j$. Additionally, we denote a vector of $M$ documents $\rmD = [\rvd_1, \ldots, \rvd_M] \in \sD^M$, and the set of $M$ combinations of documents as $\sD^{(M)}$, which contains $N^{M}$ document vectors. The marginal likelihood is defined as follows:
\begin{equation}\label{eq:mc-likelihood}
    \pth(\rva_\star | \rmQ) \defeq \sum_{\rmD \in \sD^{(M)}} \pth(\rmD | \rmQ)\, \pth(\rva_\star | \rmD, \rmQ) \,.
\end{equation} 

To model this problem, we introduce i) a reader model $\gth: \Omega^2 \rightarrow \sR$, which evaluates the likelihood of answer option $j \in [1, ..., M]$ given the query and a tuple of $K$ documents $\mathbf{d}_1, \ldots, \mathbf{d}_K$, and ii) we define a truncated retriever model $\pth(\rvd| \rvq_j)$ and $\rphi(\rvd| \rvq_j)$, which retrieves $K$ document specific to each answer option. As described in eq. \refp{eq:retriever-param}, these models are parameterized by scores $\fth(\rvd, \rvq_j)$ and $\fphi(\rvd, \rvq_j)$ respectively. The reader and retriever models are defined as:
%
\begin{gather}\label{eq:mc-reader-rertriever-model}
\pth(\rva_\star | \rmD, \rmQ) \defeq \frac{\exp \gth(\rvd_\star, \rvq_\star)}{\sum^M_{j = 1} \exp \gth(\rvd_j, \rvq_j)} \\ 
\pth(\rmD | \rmQ) \defeq \prod_{j=1}^M \pth(\rvd_j | \rvq_j), \  \rphi(\rmD | \rmQ) = \prod_{j=1}^M \rphi(\rvd_j | \rvq_j) \,. \nonumber
\end{gather}
%

The VOD objective can be applied to approximate the marginal likelihood $\pth(\rva_\star | \rmQ)$ defined in eq. \refp{eq:mc-likelihood}. In practice, the VOD objective in a multiple-choice setting implies the retrieval of $KM$ documents per query, resulting in a conditional answering likelihood that encompasses $K^M$ unique combinations. For further details, refer to Appendix \ref{apdx:applications-mc-qa}.

\subsection{Experimental Setup}

We implement a DPR-like dual-encoder architecture for the retriever with a shared backbone and implement the multiple-choice reader following \citet{Devlin2018-qr}. We use the domain-specific BioLinkBERT~\cite{Yasunaga2022-gl} as the backbone for both models and use the MedWiki corpus for all QA experiments. This results in a total of 2$\times$110M=220M parameters; a small retrieval-augmented language model. All experiments were conducted on a single node of 8 RTX 5000 GPUs using half-precision. Further details can be found in Appendix \ref{apdx:implementation}.

\paragraph{Hybrid approximate posterior} We parameterize the score $\fphi$ of the sampling distribution using a composite BM25 score combined to a checkpoint of the retriever score $\fth$ denoted $\fphi^{\rm{ckpt}}$. Specifically, we sample documents using: 
\begin{align}
    \fphi(\rva, \rvq, \rvd) \defeq& \fphi^{\rm{ckpt}}(\rvd, [\rvq; \rva])  \\
    +& \tau^{-1} \left( \rm{BM25}(\rvq, \rvd) + \beta \cdot \rm{BM25}(\rva, \rvd) \right) \nonumber \,.
\end{align}
%
where $\tau=5$ and $\beta$ is a parameter scaled proportionally to the ratio of question and answer lengths $\nicefrac{L_\rvq}{L_\rva}$ to ensure that the BM25 score of the question does not outweigh the answer score. We use $\beta = 1 + 0.5\ \rm{max}\left\{0, \log \left( \nicefrac{L_\rvq}{L_\rva}\right) \right\}$.\footnote{We picked the parameters to target a relatively high sampling entropy, no extensive hyperparameter search was performed.} At initialization $\fth$ is uninformative, we thus set $\fphi^{\rm{ckpt}}=0$. The combination of the two scores may provide a more robust sampling distribution by utilizing both the previously learned information and secondly the BM25 relevance of the query to the document.


\begin{figure}[t]
\vskip 0.2in
\vspace{-3mm}
\begin{center}
\centerline{\includegraphics[width=\columnwidth]{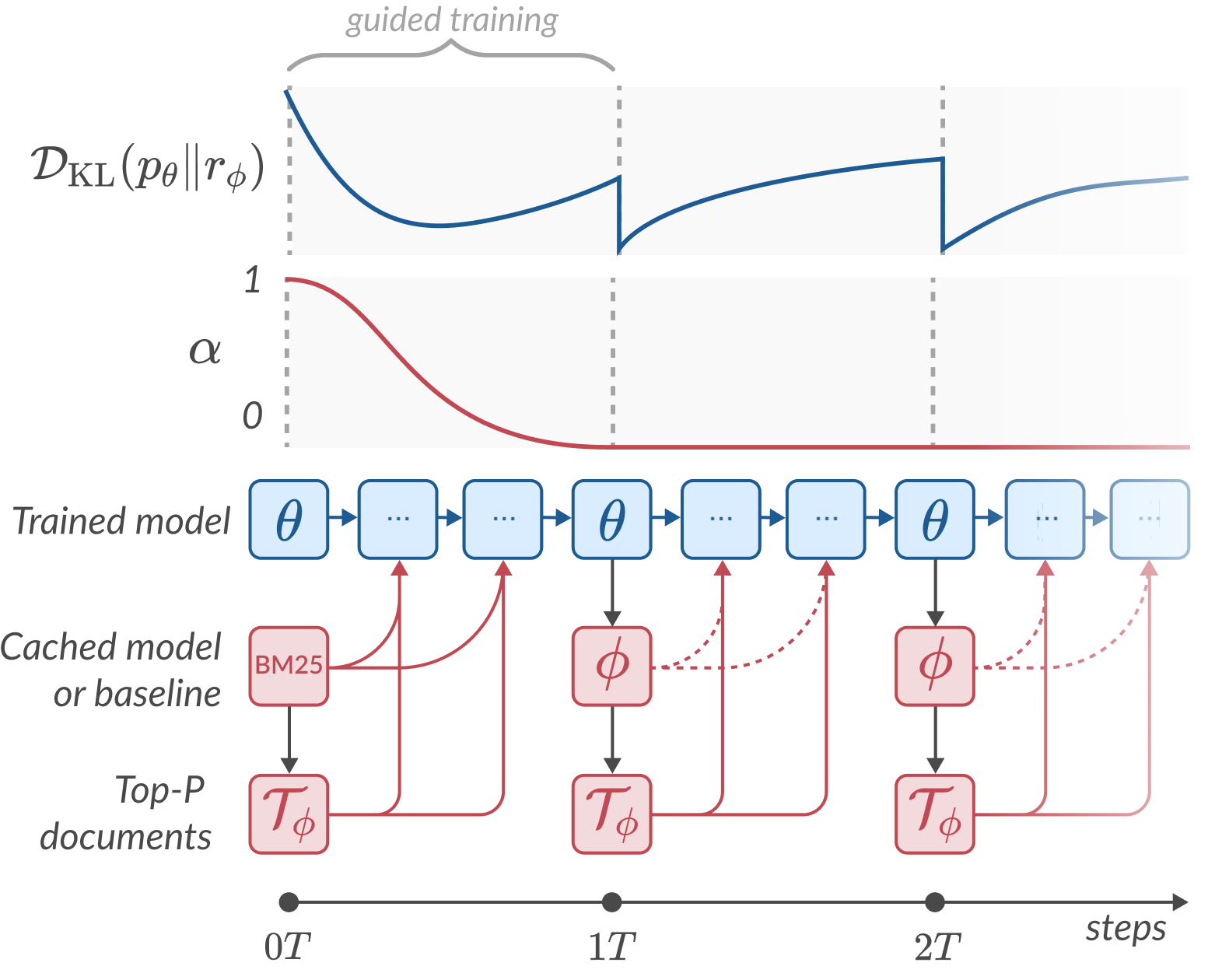}}
\caption{During training, VOD incorporates periodic updates of the cached models. In the initial period, the sampling distribution $\rphi(\rvd | \rva, \rvq)$ can be chosen as a domain-specific baseline (BM25). Additionally, a parameter $\alpha > 0$ can be utilized to guide the optimization of $\theta$. Note that the approximations $\hat{L}_{\alpha=1}^K \approx \mathcal{L}_\mathrm{ELBO}$ and $\hat{L}_{\alpha=0}^K \approx \log \pth$ can be observed, demonstrated in the experimental curves in Appendix \ref{apdx:experimental-results}.}
\label{fig:periodic-training}
\end{center}
\vskip -0.1in
\end{figure}

\paragraph{Training, periodic re-indexing and annealing} We organize the training into rounds of $T$ steps similarly to \citet{Khattab2021-xu}. As the model is exposed to a progressively larger portion of the dataset over multiple rounds, we expect optimization will result in improved generalization capabilities. At the beginning of each round, for each question-answer pair $\rvq_j$, we retrieve the set of top-$P$ documents $\gT_\phi$ and cache the set of values $\{\fphi(\rva_j, \rvq, \rvd) \mid \rvd \in \gT_\phi \}$, except for the first period where $\fphi^{\rm{ckpt}}$ is set to zero. 
During the first round, we anneal the RVB parameter $\alpha$ from 1 to 0 to stabilize early training by distilling the BM25 cached score $\fphi(\rva, \rvd, \rvq ) = 0 + \tau^{-1} \left( \rm{BM25}(\rvq, \rvd) + \beta \cdot \rm{BM25}(\rva, \rvd) \right)$ into the trainable retriever score $\fth(\rvd, \rvq)$, as shown in Figure \ref{fig:periodic-training}. At each training iteration, we sample a set of $K=8$ document $\gT_\phi$ for each of the $M=4$ question-answer pairs and evaluated the VOD objective and its gradient using the cached values of $\fphi(\rva_j, \rvq, \rvd)$.



\paragraph{Evaluation}

At evaluation time, we estimate the likelihood for each answer option using $C=10$ Monte-Carlo samples, each containing $M K = 4 \cdot 8 = 32$ documents using the estimates defined in eq. \refp{eq:vod-objective} (see Appendix \ref{apdx:applications-mc-qa}). Leveraging more samples at inference time allows for approximating the answer likelihood more robustly, as it allows for testing a greater number of combinations of documents.


\begin{table}[t]
\vspace{-1mm}
\caption{Open-domain question answering accuracy.}
\vspace{-2mm}
\label{tab:odqa-accuracy}
\vskip 0.15in
\begin{center}
\begin{small}
\resizebox{\columnwidth}{!}{%
  \begin{tabular}{lccrrrr}
    \toprule
    &&& \multicolumn{2}{c}{\centering \textbf{MedMCQA}} & \multicolumn{2}{c}{\centering \textbf{USMLE}} \\
    \bf Method & \bf Params.& \bf Finetuning & \bf Valid. & \bf Test & \bf Valid. & \bf Test \\
    \midrule
    \textbf{VOD} {\scriptsize BioLinkBERT+BM25} & 110M & MedMCQA & 51.6 & 55.3 & -- & -- \\
    \textbf{VOD} {\scriptsize BioLinkBERT+BM25} & 110M & USMLE & -- & -- & 41.0 & 40.4 \\
    \textbf{VOD} {\scriptsize 2$\times$BioLinkBERT}  & 220M & MedMCQA & 58.3 & \textbf{\underline{62.9}} &  47.2 & 46.8 \\
    \textbf{VOD} {\scriptsize 2$\times$BioLinkBERT} & 220M & USMLE & -- & -- & 45.8 & 44.7 \\
    \textbf{VOD} {\scriptsize 2$\times$BioLinkBERT}  & 220M & \tiny{MedMCQA$\rightarrow$USMLE}\textsuperscript{$\star$} & -- & -- &  53.6 & 55.0 \\
    \midrule
    \textbf{Disjoint} {\scriptsize PubMedBERT+DPR}\textsuperscript{1} & 220M & MedMCQA & 43.0 & 47.0 & -- & -- \\
    \textbf{Disjoint} {\scriptsize PubMedBERT+BM25}\textsuperscript{2} & 110M & USMLE & -- & -- & -- & 38.1 \\
    \textbf{Disjoint} {\scriptsize BioLinkBERT+BM25}\textsuperscript{3} & 110M & USMLE & -- & -- & -- & 40.0 \\
    \textbf{Disjoint} {\scriptsize BioLinkBERT-L+BM25}\textsuperscript{3} & 340M & USMLE &&& -- & 44.6 \\
    \midrule
    \textbf{Reader only} {\scriptsize PubMedGPT}\textsuperscript{4} & 2.7B &\tiny{MedMCQA+USMLE} & -- & 50.3 & -- & -- \\
    \textbf{Reader only} {\scriptsize Galactica}\textsuperscript{5} & 120B& MedMCQA & 52.9 & -- & -- & 44.4\\
    \textbf{Reader only} {\scriptsize Codex 5-shot CoT}\textsuperscript{6} & 175B & $\emptyset$ & 59.7 & 62.7 & -- & 60.2 \\
    \textbf{Reader only} {\scriptsize FLAN-PaLM}\textsuperscript{7} & 540B & $\emptyset$ & -- & 56.5 & -- & 60.3 \\
    \textbf{Reader only} {\scriptsize Med-PaLM}\textsuperscript{7} & 540B & \tiny{MedMCQA+USMLE} & -- & 57.6 & -- & \textbf{\underline{67.6}} \\
    \midrule
    \textbf{Random} {\scriptsize Uniform} & & & 25.0& 25.0 & 25.0 & 25.0 \\
    \textbf{Human} {\scriptsize Passing score}\textsuperscript{6} & & & 50.0 & 50.0 & 60.0 & 60.0\\
    \textbf{Human} {\scriptsize Merit candidate}\textsuperscript{6} & & & 90.0 & 90.0 & 87.0 & 87.0\\
    \bottomrule
    \addlinespace[0.1cm]
    \multicolumn{7}{l}{\small
        \textsuperscript{1}results from \citet{Pal2022-ph}, model from \citet{Gu2021-jn},
        ~\textsuperscript{2}\citet{Gu2021-jn}
    }\\
    \multicolumn{7}{l}{\small
        \textsuperscript{3}\citet{Yasunaga2022-gl},
        ~\textsuperscript{4}\citet{Venigalla2022-dd},
        ~\textsuperscript{5}\citet{Taylor2022-ws},
        ~\textsuperscript{6}\citet{Lievin2022-an}
    }\\
    \multicolumn{7}{l}{\small
        \textsuperscript{7}\citet{Singhal2022-la},
        ~\textsuperscript{$\star$}First pretrained on MedMCQA then finetuned on the USMLE
    }
  \end{tabular}
  }
\end{small}
\end{center}
\vskip -0.1in
\end{table}

\subsection{QA Benchmark}\label{sec:qa-benchmark}

\paragraph{MedMCQA} We report the validation and test accuracy of the VOD framework applied to BioLinkBERT (base) and the baselines in Table \ref{tab:odqa-accuracy}. 

VOD outperforms both the disjoint BERT-based methods and the recent Med-PaLM (540B parameters) with a new state-of-the-art test accuracy of 62.9\%, +0.2\% over Codex 5-shot CoT. This is an improvement of +5.3\% over Med-PaLM despite using 2.500$\times$ fewer parameters. VOD scored +7.6\% improvement over the BioLinkBERT reader with static BM25 retriever, and +15.9\% over the PubMedBERT reader coupled with a DPR retriever.


\paragraph{MedQA-USMLE} The validation and test accuracy are shown in Table \ref{tab:odqa-accuracy}. We found that using VOD with a BioLinkBERT backbone outperforms a BioLinkBERT reader coupled with a BM25 retriever, even when using the larger version of BioLinkBERT (44.7\% for VOD, 40.0\% for disjoint BioLinkBERT, 44.6\% for the disjoint large BioLinkBERT).

Due to the small size of MedQA-USMLE, pretraining on the MedMCQA proved beneficial. MedMCQA pretraining with USMLE fine-tuning resulted in VOD achieving a 55.0\% test accuracy, +10.4\% improvement over a large BioLinkBERT model with a BM25 retriever. However, Med-PaLM scores +12.6\% higher accuracy over the best VOD model. 

\paragraph{MMLU} Table \ref{tab:zeroshot-accuracy} compares the zero-shot performance of VOD, GPT-3, and Unified QA in the subcategories of \textit{psychology}, \textit{biology}, and \textit{health}. We reused the BioLinkBERT VOD model trained on MedMCQA only. VOD achieved an average accuracy of 54.8\% across all 12 tasks, surpassing both GPT-3 (47.0\%) and Unified QA (48.7\%). Particularly, VOD excelled in \textit{medical\_genetics} (+36.0\%), \textit{professional\_medicine} (+14.4\%), and \textit{anatomy} (+12.5\%). Although GPT-3 and Unified QA showed competitive results in certain areas, VOD's higher accuracy highlights its robustness to a wider set of medical tasks. 

\begin{table}[t]
\vspace{-1mm}
\caption{Zero-shot accuracy on MMLU (\%).}
\vspace{-2mm}
\label{tab:zeroshot-accuracy}
\vskip 0.15in
\begin{center}
\begin{small}
\resizebox{\columnwidth}{!}{%
  \begin{tabular}{llrrr}
\toprule
\textbf{Task}            & \textbf{Subcategory} & \textbf{Unified QA} & \textbf{GPT-3} & \textbf{VOD}  \\ 
\midrule
medical\_genetics & health & 40.0 & 40.0 & \textbf{\underline{76.0}} \\
high\_school\_psychology & psychology & \textbf{\underline{70.0}} & 61.0  & 60.6 \\
college\_biology & biology & 40.0 & 45.0  & \textbf{\underline{59.7}} \\
anatomy & health & 43.0 & 46.0 & \textbf{\underline{58.5}} \\
clinical\_knowledge      & health                  & 57.0                & 50.0                & \textbf{\underline{58.5}} \\
professional\_medicine   & health             & 43.0                & 38.0                & \textbf{\underline{57.4}} \\
nutrition                & health                        & 48.0                & 50.0                & \textbf{\underline{56.5}} \\
high\_school\_biology    & biology                     & 53.0                & 48.0                & \textbf{\underline{55.2}} \\
college\_medicine & health & 43.0                & \textbf{\underline{47.0}} & 46.8                \\
human\_aging & health & \textbf{\underline{55.0}} & 50.0 & 44.4                \\
virology & health & 43.0 & \textbf{\underline{44.0}} & 42.2                \\
professional\_psychology & psychology & \textbf{\underline{49.0}} & 45.0 & 42.2 \\
\midrule
\textbf{Average} & - & 48.7 & 47.0 & \textbf{\underline{54.8}} \\
\bottomrule
\end{tabular}
  }
\end{small}
\end{center}
\vskip -0.1in
\end{table}

\subsection{Ablation Study}\label{sec:ablation-study}

\begin{figure}[t]
\vskip 0.15in
\begin{center}
\centerline{\includegraphics[width=\columnwidth]{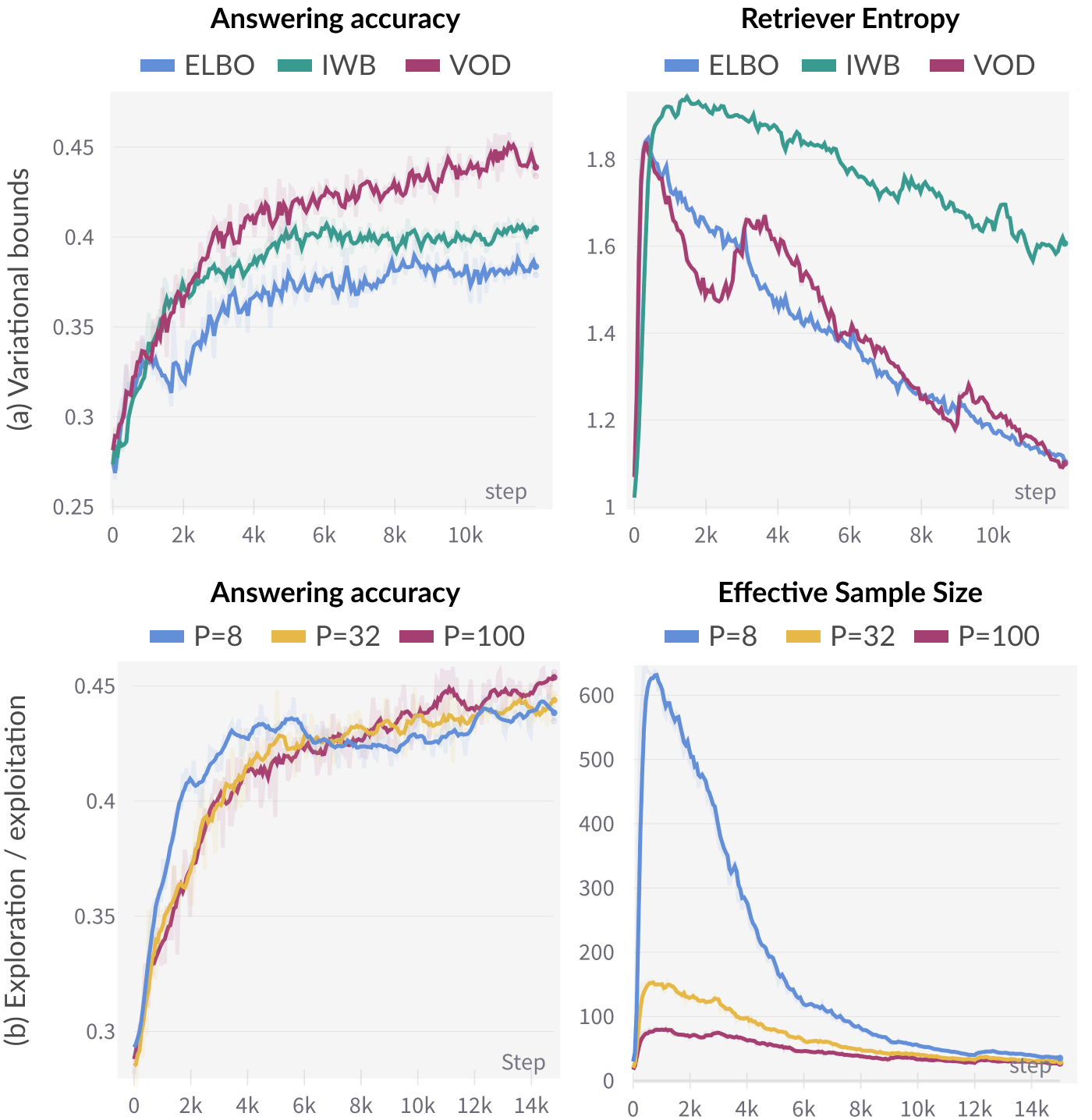}}
\caption{Answering accuracy and retriever entropy. \textbf{(a) Variational bounds:} effect of the choice of parameter $\alpha$ (ELBO: $\alpha = 1$, IWB: $\alpha=0$, RVB/VOD: interpolating $\alpha$ from 1 to 0), all using $P=100$.
\textbf{(b) Exploration / exploitation:} effect of the support size $P$ of the truncated retrievers. We sampled $MK = 4 \cdot 8$ documents per question, resulting in $K^M=4.096$ documents combinations (therefore the max. effective sample size is 4096). Higher $P$ values leads to smaller effective sample sizes, slower learning but better end performances.
 }
\label{fig:ablation}
\end{center}
\vskip -0.15in
\end{figure}

In Figure \ref{fig:ablation}, we report the performances of a VOD model for multiple variational bounds and diverse truncated retriever support sizes (the number of cached top-$P$ documents). \footnote{To reduce overall running costs, we used a dual-encoder reader with score function $\gth(\rvd, \rvq) = \bert(\rvq)^T \bert(\rvd)$.}

\paragraph{Variational bounds} We tested multiple variational bounds: the ELBO, the importance-weighted bound (IWB) and the RVB as possible methods to optimize the model. The ELBO and IWB are special cases of the RVB. For the RVB, we anneal the parameter $\alpha$, as in the main experiments, and found that this method resulted in the highest answering accuracy while also resulting in low retriever entropy. This suggests that the retriever was also optimized at a faster rate.

\paragraph{Exploration vs. Exploitation} We experimented with using values of $P \in \{8, 32, 100\}$. Using the highest value of $P=100$ resulted in a smaller effective sample size,\footnote{The effective sample size is correlated with the inverse of the variance of $\wall(\rva, \rvq, \rvd)$, it is a popular diagnostic for importance sampling. See \citet{Kong1992-mz, Owen2013-hr, Nowozin2015-eo}.} slower learning but ultimately higher accuracy.

\subsection{Information retrieval}

Despite good QA accuracy, the ability of VOD to yield a meaningful retriever component through the proposed reader-retriever end-to-end training remains, at this point of the paper, to be proven. Thus, we benchmarked a VOD retriever trained on MedMCQA against the FindZebra API\footnote{ \url{https://www.findzebra.com/api/}}, which connects to a specialized BM25 search engine targeted to medical professionals \citep{Dragusin2013-jw}. The comparison was done using the set of FindZebra queries and corpus, where searching documents using a BERT-based retriever translates into a nearest neighbour search problem in the embedding space, which we visualize in Appendix \ref{apdx:experimental-results}. 

\paragraph{Re-purposing MCQA retrievers for semantic search} The BioLinkBERT VOD model, trained on the MedMCQA dataset, has a retriever component that is trained to rank documents using question-answer pairs $[\rvq; \rva]$ as inputs (see eq. \refp{eq:mc-reader-rertriever-model}). Thus, further task adaptation is required to rank documents solely based on queries, and without answer option (e.g., using a model $\pth(\rvd | \rvq)$ instead of $\pth(\rvd | [\rvq ; \rva_j])$). To address this, we use the retriever to teach a query-only student model, which corresponds to \textit{knowledge distillation} \cite{Hinton2015-oo}. Given pairs of MedMCQA question and answers $(\rvq, \rva_\star)$, this translates into minimizing:
\begin{equation}\label{eq:answer-free-distillation}
    L_\textsc{Distill.} = \KL(\ 
    \underbrace{\rphi(\rvd \mid [\rvq; \rva_\star])}_{\text{\shortstack{MCQA Teacher \\ (question+answer)}}}
    \| 
    \underbrace{\pth(\rvd \mid \rvq)}_{\text{{\shortstack{Student \\ (question only) }}}}
    ) \,.
\end{equation}
\paragraph{Metrics} In line with \citet{Dragusin2013-jw}, we evaluate retrieval by recording the first article that matches the reference CUI (disease concept) and report 100 $\times$ the mean reciprocal rank (MRR) and the fraction of queries for which the correct article is returned in the top 20.\footnote{We re-used two of the metrics introduced in the original study. We considered the MRR to be more adequate than NDCG because not all documents with a relevant CUI can be considered as a relevant match; only the highest-ranking one is essential.}

\paragraph{Retrieval performances} We evaluated the VOD retriever with and without distillation, a hybrid retriever combining the VOD and BM25 score (defined as $\fth^{\textsc{VOD+BM25}}(\rvd, \rvq) \defeq \fth(\rvd, \rvq) + \tau^{-1}\  \textsc{BM25}(\rvd, \rvq)$ where $\tau=5$), and BM25 alone. We found that a VOD retriever trained on MedMCQA via distillation can be competitive with the FindZebra API and achieves best performances when combined with a simple BM25 baseline, resulting in an MRR of 38.9. 

\begin{table}[t]
\caption{Retrieval performances on the FindZebra benchmark for a BioLinkBERT retriever trained using VOD on MedMCQA and one trained using task-specific distillation, with and without coupling with a BM25 score during evaluation.}
\label{tab:information-retrieval-benchmark}
\vskip 0.15in
\begin{center}
\begin{small}
\begin{sc}
\begin{tabular}{lc|cc}
    \toprule
    \bf Method & \bf Distillation & \bf MRR & \bf Hit@20 \\
    \midrule
    VOD & \xmark & 27.8 & 56.9 \\
    VOD & \cmark & 31.7 & 58.1 \\
    VOD + BM25 & \cmark & \textbf{\underline{38.9}} & \textbf{\underline{64.1}} \\
    \midrule
    BM25 & -- & 26.4 & 48.4 \\
    FindZebra API & -- & 30.1 & 59.3 \\
    \bottomrule
\end{tabular}
\end{sc}
\end{small}
\end{center}
\vskip -0.1in
\end{table}

\paragraph{Retriever samples} In Appendix \ref{apdx:experimental-results}, Table \ref{tab:retrieval-samples}, we present examples of a distilled VOD retriever's top-1 ranked passages, including two successes and two failures. The top-ranked documents were mostly relevant, but the retriever struggled with long keyword-based queries, as shown in row \#4. This is likely due to the discrepancy of tasks between training on MedMCQA and evaluating on FZ queries.

\section{Discussion}

\paragraph{Knowledge vs. Reasoning Tasks} The VOD framework was evaluated using the MedMCQA and USMLE datasets only utilizing BERT-based models. The MedMCQA dataset is designed to evaluate the knowledge of entry-level medical students, whereas the USMLE dataset targets trained medical professionals, who are expected to possess not only a comprehensive understanding of medicine but also the ability to reason about complex medical problems. The results obtained demonstrate the effectiveness of the VOD framework in the specific tasks, however, we speculate that a BERT-sized model may not be sufficient for handling reasoning-intensive questions. As reported in previous studies, larger models like PaLM and Codex, have shown exceptional performance in handling reasoning-heavy questions \cite{Singhal2022-la, Lievin2022-an}. 

\paragraph{Large-scale datasets}  The nature of the task is not the sole factor limiting the performance of VOD. We showed that an initial round of training on the larger MedMCQA dataset (182k samples) strongly benefit performances on the USMLE dataset (10k samples) .\footnote{Pretraining on MedMCQA improved downstream USMLE accuracy by +10.3\% when compared to training on USMLE only.} This suggests that VOD might benefit from larger-scale training, including other tasks such as retrieval-augmented language modelling.

\paragraph{Importance sampling}

In contrast to other methods, VOD requires defining the sampling distribution explicitly and thus makes the diagnosis of the suitability of the sampling distribution possible. As utilized in Figure \ref{fig:ablation}, we suggest relying on the effective sample size diagnostic to measure the robustness of the likelihood estimates. A small effective sample size, with a value close to one, hints at a mismatch between the sampling distribution $\rphi(\rvd | \rva, \rvq)$ and the posterior $\pth(\rvd | \rva, \rvq)$. In that case, the sampling distribution should be adapted and/or optimized end-to-end with the model. Furthermore, the $\alpha$ parameter of the VOD objective can be increased towards one to target looser variational bounds, which often come with a better optimization profile \cite{Rainforth2018-pw}.

\paragraph{Approximating the IW-RVB}


The VOD objective serves as an approximate estimation of the IW-RVB, although its approximation error remains unaddressed. While the VOD objective is consistent w.r.t. the RVB (Appendix \ref{apdx:vod-objective}), its reliance on the self-normalization introduces a deviation from the strict guarantee of being a lower bound for the marginal log-likelihood, which is provided by the IW-RVB. Nonetheless, the utilization of self-normalized importance sampling is generally preferred over un-normalized approaches due to its ability to reduce variance. To thoroughly understand the bias of the VOD objective and its gradient, additional theoretical analysis is required. Despite this, the VOD objective has demonstrated sufficient robustness in enabling end-to-end training of retrieval-augmented systems and efficiently bridging the performance gap that remained with larger, non-retrieval-augmented language models, as shown in Figure \ref{fig:vod-benchmark}.

\section{Conclusion}
In conclusion, this study has provided a comprehensive examination of methods for enhancing retrieval-augmented models through variational inference. The proposed probabilistic framework, VOD, is a promising solution for achieving tractable, consistent, and end-to-end training of retrieval-augmented models. Through a series of extensive experiments on multiple-choice medical exam questions, utilizing the MedMCQA and MedQA-USMLE datasets, the effectiveness of the proposed framework have been demonstrated. The findings indicate that leveraging the Rényi variational bound yields better end-to-end performances while also optimizing at a faster rate. Additionally, this study has introduced truncated retriever parameterization with variable support size $P$, which generalizes existing top-$K$ parameterization and allows for likelihood-based optimization based on the full range of documents. Furthermore, the results have shown that VOD outperforms the state-of-the-art Codex and domain-tuned Med-PaLM on MedMCQA in terms of both accuracy and parameter efficiency.

In the future, we plan to investigate various variations of VOD to enhance its versatility in modeling other datasets and tasks, as well as exploring the possibility of jointly learning the approximate posterior. Overall, this research provides a promising direction for designing and training likelihood-based models for retrieval-augmented tasks. We hope this research will help popularizing recent advances in variational inference and importance sampling, in the field of natural language processing and beyond.

\section*{Acknowledgements}
VL's work was funded in part by Google DeepMind through a PhD grant.
OW’s work was funded in part by the Novo Nordisk Foundation through the Center for Basic Machine Learning Research in Life Science (NNF20OC0062606).
VL and OW acknowledge support from the Pioneer Centre for AI, DNRF grant number P1. 

\vfill

\bibliography{main}
\bibliographystyle{icml2023}

\newpage
\appendix
\onecolumn

\section{Priority sampling}\label{apdx:priority-sampling}

\begin{figure}[h]
\begin{center}
\includegraphics[width=\columnwidth]{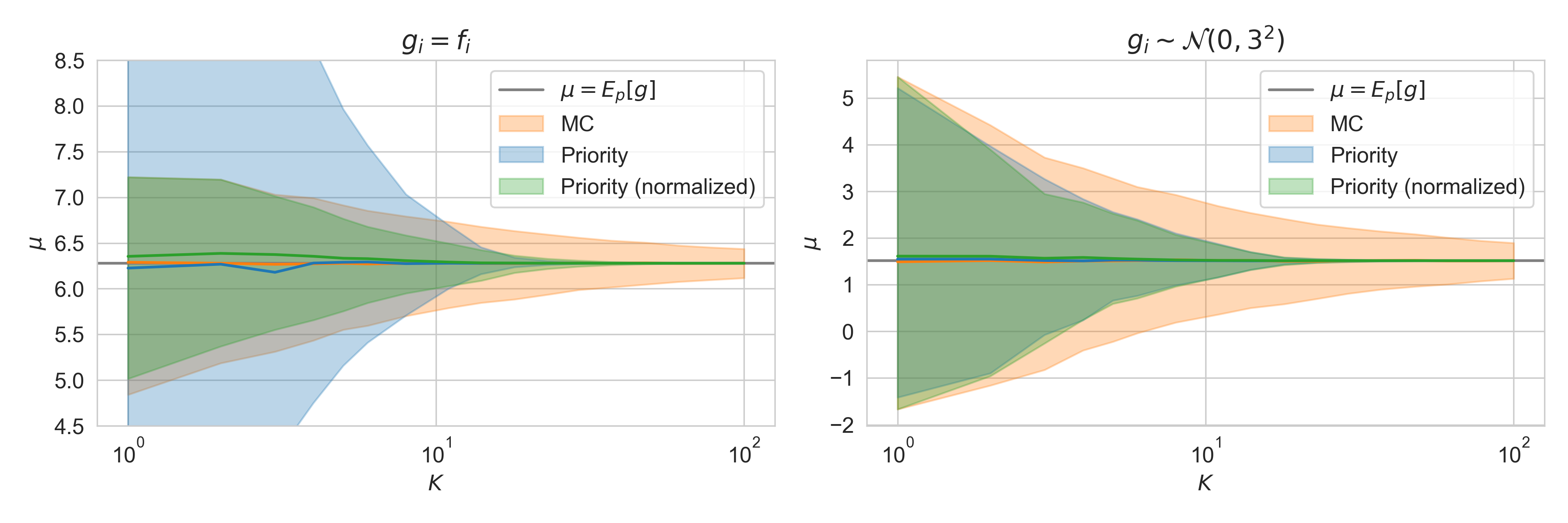}
\end{center}
\caption{Estimation of the weighted average $\mu=\Ex_p[g]$ with weights $p_i \defeq \sum_{i=1}^N \nicefrac{\exp {f_i}}{\sum_{j=1} \exp {f_j}}$ where $f_i \sim \mathcal{N}(0, 3^2)$ and $N=100$. We compare standard Monte-Carlo (sampling with replacement) with priority sampling and with self-normalized priority sampling (sampling without replacement). In the \textbf{left side} of the plot, we use $g_i = f_i$. In the \textbf{right side}, we use independent values $g_i \sim \mathcal{N}(0, 3^2)$ (sampled independently of $f_i$). We report the 80\% CI interval for 10k estimates, each with $K=1\ldots100$. Priority sampling achieves higher variance than standard MC when $g_i=f_i$. Self-normalized priority sampling achieves lower variance than standard MC.}
\label{fig:priority}
\end{figure}

Given a set of probabilities $p_1, \dots, p_N$ and a function with values $f_1, \dots, f_N$, priority sampling \cite{Duffield2007-rd} allows estimating the sum $\sum_{i=1}^N p_i f_i$ using a subset of $K < N$ samples drawn \textit{without replacement}.\footnote{We recommend \citet{Vieira2017-yd} for a great introduction to priority sampling.} For a sequence of random weights $u_1, \ldots, u_{n} \iidsim \mathrm{Uniform}(0,1]$, we define the priority keys $p_i / u_i$, set $\tau$ to be the $K+1$-th largest key, and define the set of $K$ samples $\sS = \{ i \in [1, N] \ |\ p_i / u_i > \tau \}$. Using importance-weights $\bar{s}_i := \max(p_i,\tau)$, priority sampling yields an unbiased estimate of the weighted mean:
\begin{equation}\label{eq:priority}
    \Ex_{p(u_1,\ldots,u_N)}\left[\sum_{i\in \sS} \bar{s}_i f_i\right] = \sum_{i=1}^N p_i f_i \ .
\end{equation}
\paragraph{Self-normalized importance sampling} Empirically, the estimator eq.~\refp{eq:priority} might suffer from high variance. We follow \cite{Kool2019-qp} and use self-normalize importance weights defined as $s_i \defeq \nicefrac{\bar{s}_i}{\sum_{j \in \sS} \bar{s}_j}$ to reduce variance at the cost of introducing a bias. However, the estimator $\sum_{i\in \sS} s_i f_i$ is biased but consistent: it equals the true expected value for $K=N$. The VOD objective uses self-normalized priority sampling.

\paragraph{Illustration} In Figure \ref{fig:priority}, we visualize the variance of a standard Monte-Carlo (MC) estimator in two cases, a priority sampling estimator and a priority sampling estimator with self-normalized weights. In both cases, the variance of the self-normalized priority estimate is upper-bounded by the variance of the standard MC estimate and converges to zero at a faster rate than the traditional MC estimator. In one of the two cases, the un-normalized priority estimator suffers from large variance whereas the self-normalized priority estimator benefits from lower variance in both cases.

\paragraph{Product of priority sampling estimates} 

Let $\rmZ = [\rvz_1, \ldots, \rvz_M]$ be a vector of $M$ independent variables, each defined on sets $\sZ_1, \ldots, \sZ_M$, each of size $N$. The vector $\rmZ$ is defined on the set $\sZ^{(M)} = \sZ_1 \times \ldots \times \sZ_M$, the Cartesian product of the $M$ sets, which corresponds to $N^M$ combinations. Given a probability distribution $p(\rmZ) = \prod_{j=1}^M p(\rvz_j)$, we draw $K$ samples for each component using priority sampling:
\begin{subequations}\label{eq:per-option-sampling}
\begin{gather}
\sS_j = \{\rvz_{j,1}, \ldots, \rvz_{j,K} \}  \\
(\rvz_{j,1} , s_j[\rvz_1]), \ldots, (\rvz_{j,K}, s_j[\rvz_K]) \prioritysim p(\rvz_j) \,.
\end{gather}
\end{subequations}

Combining the per-component priority samples $p(\rmZ | \rmQ)$ by defining the product priority weight allows estimating an average of a function $h(\rmZ)$ weighted by $p(\rmZ)$. Defining the product of priority weights as $s(\rmZ) \defeq \prod_{j=1}^{M} s_j[\rvz_j]$, we have:
\begin{subequations}
\begin{align}
    \Ex_{p(\rmZ)} \left[ h(\rmZ) \right] =& \Ex_{p(\rvz_1))} \left[ \ldots  \left[
    \Ex_{p(\rvz_M)} \left[  h(\rmZ) \right] \right] \ldots \right] \\ 
    \approx& \sum_{\rvz_1 \in \sS_1} s_1[\rvz_1] \ldots  \sum_{\rvz_M \in \sS_M} s_M[\rvz_M] h(\rmZ) \\
    = & \sum_{\rvz_1 \in \sS_1} \ldots  \sum_{\rvz_M \in \sS_M} s_1[\rvz_1] \ldots s_M[\rvz_M] h(\rmZ) \\
    =& \sum_{\rmZ \in \sS^{(M)}} s(\rmZ) h(\rmZ) \ .
\end{align}
\end{subequations}

\section{VOD objective}\label{apdx:vod-objective}

\begin{tcolorbox}[breakable,colback=lightgray,colframe=darkgray,left=4pt,right=4pt,top=4pt,bottom=4pt]

Given a reader model $\pth(\rva | \rvd, \vq)$, and retriever model $\pth(\rvd | \rvq)$ and a proposal $\rphi(\rvd | \rva, \rvq)$, the VOD objective is:
\begin{subequations}
\begin{gather}\label{eq:apdx-vod-objective}
\hat{L}^{K}_{\alpha}(\rva, \rvq) \defeq
    \frac{1}{1-\alpha} \log \sum_{i=1}^K s_i\, \hat{v}_{\theta,\phi}^{1-\alpha} (\rva, \rvq, \rvd_i) \\
    (\rvd_1, s_i), \ldots, (\rvd_K, s_K) \prioritysim \rphi(\rvd | \rva, \rvq) \\
    \hat{v}_{\theta,\phi} \defeq\, \pth(\rva| \rvq, \rvd_i) \zeta (\rvd_i) \left(\sum_{j =1}^K s_j \zeta(\rvd_j)\right)^{-1} \,.
\end{gather}%
\end{subequations}
The VOD objective is a self-normalized importance sampling estimate of the RVB, and thus converges with probability one (\textit{consistency}). Denoting $\gT_\phi$ the support of $\pth(\rvd|\rvq)$, we have: 
\begin{equation}
    \lim_{K \rightarrow |\gT_\phi|} \underbrace{\hat{L}_\alpha^K (\rva, \rvq)}_{\text{VOD}}
    =
    \underbrace{\mathcal{L}_\alpha(\rvd, \rvq)}_{\text{RVB}} \,.
\end{equation}

\end{tcolorbox}

Without loss of generality, we consider a joint reader-retriever model $\pth(\rva, \rvd | \rvq) = \pth(\rva| \rvd, \rvq) \pth(\rvd|\rvq)$ with retriever and sampling distribution defined on a support of documents $\gT_\phi$\footnote{$\gT_\phi$ can be chosen as the entire corpus of documents.} and parameterized as
\begin{align}\label{eq:apdx-retrievers-param}
    \pth(\rvd | \rvq) \defeq& Z_\theta^{-1} \exp \fth(\rvd, \rvq), \quad &\rphi(\rvd | \rva, \rvq) \defeq& Z_\phi^{-1} \exp \fphi(\rva, \rvd, \rvq) \\
    Z_\theta \defeq& \sum_{\rvd \in \gT_\phi} \exp \fth(\rvd, \rvq), \quad&Z_\phi \defeq& \sum_{\rvd \in \gT_\phi} \exp \fphi(\rva, \rvd, \rvq) \,.
\end{align}

In this section, we first detail the properties of the VOD objective: its complexity and its relation to the importance-weighted Rényi variational bound (IW-RVB). As a second step, we derive the VOD objective and prove that it is consistent: the VOD objective converges to the IW-RVB with probability 1 as $K \rightarrow \infty$.

\subsection{Complexity $\mathcal{O}(K)$} Evaluating the VOD objective eq. \refp{eq:apdx-vod-objective} only requires evaluating $\pth(\rva | \rvd, \rvq)$ (complexity $\mathcal{O}(1)$, generally one BERT/LM call) and evaluating the retrieval score $\fth(\rvd, \rvq)$ for each document $\rvd_1, \ldots, \rvd_K$ (complexity $\mathcal{O}(1+K)$, generally one BERT/LM call per document and one call to encode the query $\rvq$). \textit{\underline{Evaluating the VOD objective does not require evaluating the constant $Z_\theta$}} (complexity $\mathcal{O}(P)$ , one call for each document in the set $\gT_\phi$). This results in a computational complexity of $\mathcal{O}(2+K) = \mathcal{O}(K)$.\footnote{The scores $\fphi(\rvd_1), \ldots, \fphi(\rvd_K)$ of the sampling distribution are computed offline and therefore can be ignored.}

\subsection{VOD, IW-RVB, ELBO and marginal likelihood}

Using a set $\rvd_1, \ldots, \rvd_K \sim \rphi(\rvd | \rva, \rvq)$ sampled with replacement, the importance-weighted Rényi variational bound (IW-RVB) is defined as:
\begin{equation}\label{eq:apdx-iw-rvb}
\hat{\mathcal{L}}^K_\alpha(\rvd, \rvq) \defeq \frac{1}{1 - \alpha} \log \frac{1}{K} \sum_{i=1}^{K} \wall^{1-\alpha}(\rva, \rvq, \rvd_i) \,.
\end{equation}
The IW-RVB is a lower-bound of the log-likelihood and for $\alpha=0$, increasing the number of samples results in a tighter log-likelihood lower bound \cite{Burda2015-wt}:
\begin{equation}
    \VIB(\rva, \rvq) 
    \leq \hat{\mathcal{L}}^{K}_{\alpha=0}(\rvd, \rvq) 
    \leq \hat{\mathcal{L}}^{K+1}_{\alpha=0} (\rvd, \rvq) 
    \leq \log \pth(\rva, \rvq) 
    \,.
\end{equation}
In $\alpha=0$, the RVB is defined by continuity as the ELBO \cite{Li2016-ph}. In that case, increasing the number of Monte Carlo samples $K$ does not result in a tighter bound:
\begin{equation}
    \hat{\mathcal{L}}^{K}_{\alpha \rightarrow 1}(\rvd, \rvq) 
    = \Ex_{\rphi(\rvd_1, \ldots, \rvd_K | \rva, \rvq)} \left[  \frac{1}{K} \sum_{i=1}^K \log \wall(\rva, \rvq, \rvd_i) \right]
    = \Ex_{\rphi(\rvd | \rva, \rvq)} \left[ \log \wall(\rvq, \rva, \rvd) \right]
    = \VIB(\rva, \rvq) 
    \,.
\end{equation}

The VOD objective is a self-normalized importance sampling estimate of the RVB, whereas the IW-RVB is a standard importance sampling. The VOD objective only differs from the IW-RVB because (i) VOD relies on self-normalized priority sampling eq. \refp{eq:apdx-vod-vs-iwrvb-priority}, (ii) the normalizing constant $Z_\theta Z_\phi^{-1}$ in the expression of the importance weight $\wall(\rva, \rvq, \rvd)$ is estimated with a self-normalized priority sampling estimate eq. \refp{eq:apdx-vod-vs-iwrvb-weight}.

\subsection{Derivation of the VOD objective}

In this section, we derive the VOD objective. We begin by expressing the ratio of normalization constants $\nicefrac{Z_\theta}{Z_\phi}$ as a function of $\zeta$ (section \ref{apdx-sec:ratio}), and then apply this identity to approximate the importance weight $\wall(\rvq, \rva, \rvd)$ (section \ref{apdx-sec:approx-iw}). We conclude the deriving the VOD objective: an approximation of the IW-RVB using (i) priority sampling and (ii) the importance weight estimate (\ref{apdx-sec:vod-objective-derivation}).

\subsubsection{Ratio of normalizing constants $\nicefrac{Z_\theta}{Z_\phi}$}\label{apdx-sec:ratio}

The quantity $\nicefrac{Z_\theta}{Z_\phi}$ can expressed as a function of the ratio of un-normalized retriever densities $\zeta(\rvd) \defeq {\exp \fth(\rvd, \rvq)} / {\exp \fphi(\rva, \rvd, \rvq)}$ using the following identity:
\begin{equation}\label{eq:apdx-ratio}
Z_\theta Z_\phi^{-1} = \Ex_{\rphi(\rvd|\rva, \rvq)} \left[ \zeta(\rvd) \right] \ .
\end{equation}

\paragraph{Proof}
 The equality arises from the definition of the right-hand term:
\begin{subequations}
\begin{align}
    \Ex_{\rphi(\rvd|\rva, \rvq)} \left[ \zeta(\rvd) \right] \defeq & \sum_{\rvd \in \gT_\phi} \rphi(\rvd | \rva, \rvq) \frac{\exp \fth(\rvd, \rvq)}{\exp \fphi(\rva, \rvd, \rvq)} \\
    = &  \sum_{\rvd \in \gT_\phi}  \frac{{\exp \fphi(\rva, \rvd, \rvq)}}{Z_\phi} \frac{\exp \fth(\rvd, \rvq)}{{\exp \fphi(\rva, \rvd, \rvq)}}
    = Z_\theta Z_\phi^{-1} \,.
\end{align}
\end{subequations}

\subsubsection{Estimation of the importance weight $\wall$}\label{apdx-sec:approx-iw}

The importance weight $\wall(\rvq, \rva, \rvd)$ can be approximated using $K$ retrieval scores $\fth(\rvd_1), \ldots, \fth(\rvd_K)$:
 \begin{subequations}
 \begin{gather}
    \wall(\rvq, \rva, \rvd) \approx \hat{v}_{\theta,\phi}(\rvq, \rva, \rvd) \defeq \pth(\rva| \rvq, \rvd) \zeta (\rvd) \left(\sum_{j =1}^K s_j \zeta(\rvd_j)\right)^{-1}
     \label{eq:apdx-iw-approximate}\\
    (\rvd_1, s_i), \ldots, (\rvd_K, s_K) \prioritysim \rphi(\rvd | \rva, \rvq) \nonumber \,.
\end{gather}
\end{subequations}

\paragraph{Proof} Using the eq. \refp{eq:apdx-ratio}, we can express $\wall(\rvq, \rva, \rvd)$ as a function of the un-normalized retriever density ratio $\zeta$:
\begin{subequations}
\begin{align}
\wall(\rva, \rvd, \rvq) \defeq & \frac{\pth(\rva | \rvd, \rvq) \pth(\rvd | \rvq)}{\rphi(\rvd | \rva, \rvq)} \\ 
= & \pth(\rva | \rvd, \rvq) \zeta(\rvd) \left( Z_\theta Z_\phi^{-1} \right)^{-1} \\
= & \pth(\rva | \rvd, \rvq) \zeta(\rvd) \left( \Ex_{\rphi(\rvd|\rva, \rvq)} \left[ \zeta(\rvd) \right] \right)^{-1} \,.
\end{align}
\end{subequations}
The expected value of $\zeta(\rvd)$ can be estimated via Monte Carlo. Using priority sampling with samples $\rvd_1, \ldots, \rvd_K \sim \rphi(\rvd | \rva, \rvq)$ and normalized priority weights $s_1, \ldots, s_K$ (section \ref{apdx:priority-sampling}), we obtain:
\begin{align}
    \wall(\rva, \rvd, \rvq) \approx {\pth(\rva | \rvd, \rvq) \zeta(\rvd) } \left({\sum_{j=1}^K s_j \zeta(\rvd_j) }\right)^{-1}  = \vall(\rva, \rvd, \rvq) \,.
\end{align}

\subsubsection{The VOD objective}\label{apdx-sec:vod-objective-derivation}

Given document samples $\rvd_1, \ldots, \rvd_K \prioritysim \rphi(\rvd | \rva, \rva)$ with self-normalized priority weights $s_1, \ldots, s_K$. The VOD objective $\hat{L}^K_\alpha(\rvd, \rvq) $ is an approximation of the IW-RVB ($\hat{\mathcal{L}}^K_\alpha(\rvd, \rvq)$, eq. \refp{eq:apdx-iw-rvb}):
\begin{subequations}
\begin{align}
\hat{\mathcal{L}}^K_\alpha(\rvd, \rvq) \approx & \frac{1}{1 - \alpha} \log \sum_{i=1}^{K} s_i\, \wall^{1-\alpha}(\rva, \rvq, \rvd_i) \quad & \text{(priority sampling)} \label{eq:apdx-vod-vs-iwrvb-priority} \\
\approx & \frac{1}{1 - \alpha} \log \sum_{i=1}^{K} s_i\, \vall^{1-\alpha}(\rva, \rvq, \rvd_i) = \hat{L}^K_\alpha(\rvd, \rvq) \,. \quad & \text{(inserting eq. \refp{eq:apdx-iw-approximate})} \label{eq:apdx-vod-vs-iwrvb-weight}
\end{align}
\end{subequations}


\subsection{VOD consistency} 

In a nutshell, the VOD objective is biased because some normalization terms are estimated via Monte Carlo. Nevertheless, the estimates used as denominator are themselves consistent. This results in a final estimate -- the VOD objective -- which is itself consistent.

In contrast to the IW-RVB eq. \refp{eq:apdx-iw-rvb}, the VOD objective $\hat{L}_\alpha^K$ is not guaranteed to be a lower bound of the marginal log-likelihood. Nonetheless, the VOD objective and its gradient are consistent: they converge to their target expressions (RVB) in the limit of $K \rightarrow |\gT_\phi| < \infty$. 

\paragraph{Proof}
Self-normalized priority sampling is consistent. Given an arbitrary function $h$ such that $|h(\rvx)| < \infty$ and $K$ priority samples $(\rvx_1, s_1), \ldots, (\rvx_K, s_K) \prioritysim p(\rvx)$ where $\rvx \in \gX, |\gX| < \infty$:
\begin{align}\label{eq:apdx-priority-is-consistent}
    \lim_{K \rightarrow |\gX|} \sum_i s_i h(\rvx_i) = \lim_{K \rightarrow |\gX|} \sum_i \frac{\bar{s}_i}{\sum_j \bar{s}_j} h(\rvx_i) 
    =
    \Ex_{p(\rvx)} \left[ \frac{h(\rvx)}{\Ex_{p(\rvx)} \left[ 1 \right]} \right]
    = \Ex_{p(\rvx)} \left[ h(\rvx) \right] \,.
\end{align}
Assuming $|\zeta(\rvd)| < \infty$, this result implies that $\vall$ is a consistent estimate of the importance weight $\wall$:
\begin{subequations}
\begin{align}
    \lim_{K \rightarrow |\gT_\phi|} \vall(\rva, \rvq, \rvd) = & {\pth(\rva | \rvd, \rvq) \zeta(\rvd) } \left( \lim_{K \rightarrow |\gT_\phi|} {\sum_{j=1}^K s_j \zeta(\rvd_j) }\right)^{-1} \\
    = & {\pth(\rva | \rvd, \rvq) \zeta(\rvd) } \left( \Ex_{\rphi(\rvd|\rva, \rvq)} \left[ \zeta(\rvd) \right] \right)^{-1} \\
    = & {\pth(\rva | \rvd, \rvq) \zeta(\rvd) } \left( Z_\theta Z_\phi^{-1}\right)^{-1} \\
    = & \wall(\rva, \rvq, \rvd) \,.
\end{align}
\end{subequations}
The VOD objective relies on the importance weight estimates, which are themselves consistent. Therefore for $\alpha <1$:
\begin{subequations}
\begin{align}
    \lim_{K \rightarrow |\gT_\phi|} \hat{L}_\alpha^K(\rva, \rvq) = & \lim_{K \rightarrow |\gT_\phi|}  \frac{1}{1-\alpha} \log \sum_{i=1}^K s_i\, \hat{v}_{\theta,\phi}^{1-\alpha} (\rva, \rvq, \rvd_i) \\ 
    = &  \frac{1}{1-\alpha} \log \Ex_{\rphi(\rvd|\rva, \rvq)} \left[ 
    \lim_{K \rightarrow |\gT_\phi|} \hat{v}^{1-\alpha}_{\theta,\phi} (\rva, \rvq, \rvd_i) \right] \\ 
    = &  \frac{1}{1-\alpha} \log \Ex_{\rphi(\rvd|\rva, \rvq)} \left[ 
    \wall^{1-\alpha}(\rva, \rvq, \rvd)  \right] \\
    = & \mathcal{L}_\alpha(\rva, \rvq) = \lim_{K \rightarrow |\gT_\phi|} \mathcal{L}_\alpha^K(\rva, \rvq) \,.
\end{align}
\end{subequations}

\section{VOD gradient}\label{apdx:vod-gradients}
\begin{tcolorbox}[breakable,colback=lightgray,colframe=darkgray,left=4pt,right=4pt,top=4pt,bottom=4pt]

The VOD gradient w.r.t. the parameter $\theta$ corresponds to a self-normalized importance sampling estimate of the RVB gradient. It corresponds to the IW-RVB gradient derived in \cite{Li2016-ph}, except that further approximations are required to ensure the expression is tractable. The VOD gradient is expressed as
\begin{gather}
    \mu_{\theta,\alpha,K}^{\mathrm{VOD}} \defeq \sum_{i=1}^K \frac{s_i\, \left(\pth(\rva| \rvd_i, \rvq) \zeta(\rvd_i)\right)^{1-\alpha}}{\sum_{j=1}^K s_j\, \left(\pth(\rva| \rvd_j, \rvq) \zeta(\rvd_i)\right)^{1-\alpha}} \left( \nabla_\theta \log \pth(\rva|\rvd_i, \rvq) + \rvh(\rvd_i, \rvq) \right) \approx \nabla \mathcal{L}_\alpha^K(\rva, \rq) \label{eq:vod-gradient} \\
    (\rvd_1, s_i), \ldots, (\rvd_K, s_K) \prioritysim \rphi(\rvd | \rva, \rvq) \nonumber
\end{gather}
where
\begin{equation}
    \rvh(\rvd_i, \rvq) \defeq \nabla_\theta \fth(\rvd_i, \rvq) - \sum_{j=1}^K \frac{s_j\, \zeta(\rvd_j)}{\sum_{k=1}^K s_k\, \zeta(\rvd_k)} \nabla_\theta \fth(\rvd_j, \rvq) \approx \nabla_\theta \log \pth(\rvd | \rvq) \,.
\end{equation}

The VOD gradient is consistent: it converges to the exact gradient $\nabla_\theta \mathcal{L}_\alpha(\rva, \rvq)$ with probability one:
\begin{equation}
    \lim_{K \rightarrow |\gT_\phi|} \mu_{\theta,\alpha,K}^{\mathrm{VOD}} = \nabla_\theta \mathcal{L}_\alpha(\rva, \rvq) \,.
\end{equation}

The estimation of the gradient of the VOD objective w.r.t. the parameter $\phi$ will be left to future work. In all experiments included in this paper, the parameter $\phi$ is non trainable.

\end{tcolorbox}

\paragraph{Proof} Using the results from the previous section, the gradient of the RVB w.r.t the parameter $\theta$ can be estimated as:
\begin{subequations}
\begin{align}
    \nabla_\theta \RVB_{\alpha}(\rva, \rvq) \defeq & \Ex_{\rphi(\rvd | \rva, \rvq)} \left[  \widetilde{\wall^{1-\alpha}}(\rva, \rvq, \rvd) \ \nabla_\theta\log \pth(\rva, \rvd | \rvq) \right] \\ 
    = &  \Ex_{\rphi(\rvd | \rva, \rvq)} \left[ \frac{\wall^{1-\alpha}(\rva, \rvq, \rvd)}{\Ex_{\rphi(\rvd' | \rva, \rvq)} \left[ \wall^{1-\alpha}(\rva, \rvq, \rvd') \right]} \ \nabla_\theta\log \pth(\rva, \rvd | \rvq) \right] \\ 
    = & \Ex_{\rphi(\rvd | \rva, \rvq)} \left[ \frac{
    \left({\pth(\rva | \rvd, \rvq) \zeta(\rvd) Z_\theta^{-1} Z_\phi} \right)^{1-\alpha}
    }{\Ex_{\rphi(\rvd' | \rva, \rvq)} \left[ 
    \left( {\pth(\rva | \rvd', \rvq) \zeta(\rvd') }{Z_\theta^{-1} Z_\phi}\right)^{1-\alpha}
    \right]} \ \nabla_\theta\log \pth(\rva, \rvd | \rvq) \right] \\ 
    = & \sum_{\rvd \in \sS} \frac{ s(\rvd) \left( \pth(\rva | \rvd, \rvq) \zeta(\rvd)\right)^{1-\alpha}}{\sum_{\rvd' \in \sS} s(\rvd) \left( \pth(\rva | \rvd', \rvq) \zeta(\rvd)\right)^{1-\alpha}} \nabla_\theta  \log \pth(\rva, \rvd | \rvq) 
     \,.
\end{align}
\end{subequations}


Another approximation is required to estimate $\nabla_\theta \log \pth(\rva, \rvd | \rvq) = \nabla_\theta  \log \pth(\rvq | \rvd, \rvq) + \nabla_\theta  \log \pth(\rvd | \rvq)$ without paying the price of evaluating $Z_\theta$. We approximate the term $\nabla_\theta  \log \pth(\rvd | \rvq)$ using:
\begin{subequations}
\begin{align}
    \nabla_\theta \log \pth(\rvd | \rvq) =& \nabla_\theta \fth(\rvd, \rvq) - \nabla_\theta \log Z_\theta \\ 
    &= \nabla_\theta \fth(\rvd, \rvq) - \frac{\nabla_\theta Z_\theta}{ Z_\theta } \\ 
    =&  \nabla_\theta \fth(\rvd, \rvq) -  \sum_{\rvd' \in \gT_\phi} \pth(\rvd' | \rvq)  \nabla_\theta \fth(\rvd', \rvq) \\ 
    =&  \nabla_\theta \fth(\rvd, \rvq) -  \sum_{\rvd' \in \gT_\phi} \rphi(\rvd' | \rva, \rvq)  \frac{\pth(\rvd' | \rvq)}{\rphi(\rvd' | \rva, \rvq)} \nabla_\theta \fth(\rvd', \rvq) \\ 
    =&  \nabla_\theta \fth(\rvd, \rvq) -  \Ex_{\rphi(\rvd' | \rva, \rvq)} \left[  \frac{\zeta(\rvd')}{\Ex_{\rphi(\rvd'' | \rva, \rvq)} \left[ \zeta(\rvd'') \right]} \nabla_\theta \fth(\rvd', \rvq) \right] \\ 
    \approx & \nabla_\theta \fth(\rvd, \rvq) -  \sum_{i=1}^K \frac{s_i\, \zeta(\rvd_i)}{\sum_{j=1}^K s_j\, \zeta(\rvd_j)} \nabla_\theta \fth(\rvd_i, \rvq) \ .
\end{align}
\end{subequations}
This approximation is also consistent because self-normalized priority sampling is consistent (direct application of eq. \refp{eq:apdx-priority-is-consistent}).

\section{VOD and REALM}\label{apdx:realm}

Using the truncated retriever $\pth(\rvd | \rvq)$ defined on the support $\gT_\phi$ of the top-$K$=$P$ documents ranked by a cached score $\fphi$:
\begin{gather}
\pth(\rvd | \rvq) \defeq \frac{ \ind[\rvd \in \gT_\phi] \exp{\fth(\rvd, \rvq)}}{\sum_{i=1}^K \exp{\fth(\rvd_i, \rvq)}} \,.
\end{gather}
the VOD objective aligns with REALM in $\alpha=0$. This corresponds to the marginal log-likelihood truncated to the top $K$ documents (the first step is a direct application of priority sampling being consistent):
\begin{gather}
    \underbrace{\hat{L}_{\alpha=0}^{K=P} (\rva, \rvq)}_{\text{VOD}}
    = \log \sum_{i=1}^K \rphi(\rvd_i | \rva, \rvq) 
    \wall(\rva, \rvq, \rvd_i)
    = \log \sum_{i=1}^K \pth(\rvd_i, \rva | \rvq)  
    = \underbrace{\log \pth(\rva | \rvq )}_{\text{REALM}} \,.
\end{gather}

\section{Applications of the VOD framework}\label{apdx:applications}
In this section, we detail how to apply the VOD framework to the tasks of language modelling as well as extractive, generative and multiple-choice ODQA. We also detail a solution to optimizing multi-documents readers (FiD) jointly.

\subsection{Generative and extractive ODQA}
The model $\pth(\rva | \rvd, \rvq)$  a machine reading comprehension component that can be implemented either using an extractive approach, as done in the original BERT~\citep{Devlin2018-qr}, or using a generative approach~\cite{Lewis2019-uf}. Applying the VOD framework to generative and extractive ODQA simply requires plugging the likelihood of the corresponding machine reading comprehension model $\pth(\rva | \rvd, \rvq)$ in the VOD objective and gradient (equations \ref{eq:vod-objective} and \ref{eq:vod-gradient}).

\subsection{Retrieval-augmented language modelling}
We consider the variable $\rva = [\rva_1, \ldots, \rva_{T}]$ to be the sequence of tokens of length $T$ and omit the conditioning variable $\rvq$. The retriever model $\pth(\rvd_t | \rva_{<t})$ is defined on a set of documents $\sD$. We consider a left-to-right factorized reader $\pth(\rva) \defeq \prod_{t=1}^{T} \pth(\rva_t | \rva_{<t})$. This allows us to define the following retrieval-augmented language model, with one retrieved document per token:
\begin{equation}
    \pth(\rva) \defeq \prod_{t=1}^{T} \sum_{\rvd_t \in \sD} \pth(\rvd_t | \rva_{<t}) \pth(\rva_t | \rvd_t, \rva_{<t})  \ .
\end{equation}
We apply the RVB to each step $t$ using an sampling distribution $\rphi(\rvd_t | \rva)$, this results in the following lower bound:
\begin{subequations}
\begin{align}
    \log \pth(\rva) \geq & \log \prod_{t=1}^T  \RVB_\alpha(\rva_t, \rva_{<t}) \\
    = &  \frac{1}{1-\alpha} \sum_{t=1}^T \log \Ex_{\rphi(\rvd | \rva, \rvq)} \left[ \wall^{1-\alpha}(\rva_t, \rva_{<t}, \rvd_t)\right] \ .
\end{align}
\end{subequations}
The above step-wise RVB $\RVB_\alpha(\rva_t, \rva_{<t})$ can be estimated using equation \ref{eq:vod-objective}, its gradient is given in equation \ref{eq:vod-gradient}. 

\subsection{Fusion-in-Decoder (FiD)}

In this work, we considered reader models $\pth(\rva | \rvd, \rvq)$ with a single document per sample. Alternatively, models such as FiD~\citep{Izacard2020-ui} implement a reader model that allows reading multiple documents per sample. Given a set $\sS \defeq \{ \rvd_1, \ldots, \rvd_K \}$ of documents, we denote the multi-document reader $\pth(\rva | \sS, \rvq)$. Defining a distribution over the set of unique documents $p(\sS)$ with tractable sampling and density evaluation is challenging. EMDR \cite{Sachan2021-vq} optimized a multi-document reader jointly with a deep retriever. However, an auxiliary reader model $\pth(\rva | \sS, \rvq) \defeq \prod_{i=1}^K \pth(\rva| \rvd_i, \rvq)$ is used to optimize a retriever model $\pth(\sS| \rvq) \defeq \prod_{i=1}^K \pth( \rvd_i| \rvq)$. VOD can be applied by following the same strategy, and this is equivalent to optimizing a single-sample joint reader along with a multi-sample reader:
\begin{equation}\label{eq:multi-reader}
    \mu^\mathrm{VOD-FiD}_{\theta, \alpha, \sS} \defeq
    \underbrace{
    \nabla_\theta \log \pth(\rva | \sS, \rvq)
    }_{\text{\shortstack{multi-sample \\ reader likelihood}}}
     +
     \underbrace{
     \mu_{\theta,\alpha,K}^{\mathrm{VOD}}(\rva, \rvq, \sS)
     }_{\text{single-sample VOD gradient}} \ .
\end{equation}

\subsection{Multiple-choice ODQA}\label{apdx:applications-mc-qa}

\paragraph{Model} In the multiple-choice setting, a vector of $M$ answer options $\rmA \defeq [\rva_1, \ldots, \rva_M]$ is given. We denote $\rva$ the correct option and assume $\rva \in \rmA$. We define the vector of $M$ queries as $\rmQ = [\rvq_1, \ldots, \rvq_M]$ with $\rvq_j \defeq [\rvq; \rva_j]$ where $[\cdot ; \cdot]$ denotes the concatenation operator. We denote $\rmD = [\rvd_1, \ldots, \rvd_M]$ a vector of $M$ documents, one for each answer option. We adopt a truncated retriever parameterization, given a set $\gT_\phi(\rvq_j)$ of top-P documents ranked by a function $\fphi(\cdot, \rvq_j)$, for each answer option:
\begin{equation}\label{eq:apedx-per-option-retriever-param} 
\pth(\rvd | \rvq_j) \defeq \frac{ \ind[\rvd \in \gT_\phi(\rvq_j)] \exp{\fth(\rvd, \rvq_j)}}{\sum_{\rvd' \in \gT_\phi(\rvq_j)} \exp{\fth(\rvd', \rvq_j)}} \,, \quad
    \rphi(\rvd | \rvq_j) \defeq \frac{\ind[\rvd \in \gT_\phi(\rvq_j)] \exp{\fphi(\rvd, \rvq_j)}}{\sum_{\rvd' \in \gT_\phi(\rvq_j)} \exp{\fphi(\rvd, \rvq_j)}} 
\end{equation}

Using the per-option retriever models, we define the multiple-choice ODQA model as:\footnote{In this paper we omitted the dependency of $\rphi$ on the index of the correct answer $\rva_\star$, which could be used to improve learning performances.}
\begin{subequations} \label{eq:apdx-mc-model}
\begin{gather}
\pth(\rva_\star | \rmD, \rmQ) \defeq \frac{\exp \gth(\rvd_\star, \rvq_\star)}{\sum^M_{j = 1} \exp \gth(\rvd_j, \rvq_j)}\,, \\ 
    \pth(\rmD | \rmQ) \defeq  \prod_{j=1}^{M} \pth(\rvd_j | \rvq_j)\,, \quad
    \rphi(\rmD | \rmQ) \defeq \prod_{j=1}^{M} \rphi(\rvd_j | \rvq_j) \label{eq:apdx-mc-model-retrievers} \ .
\end{gather}
\end{subequations}

Denoting $F_\theta(\rmD, \rmQ) \defeq \sum_{j=1}^M \fth(\rvd_j, \rvq_j)$ and $F_\phi(\rmD, \rmQ) \defeq \sum_{j=1}^M \fphi(\rvd_j, \rvq_j)$, the equation \ref{eq:apdx-mc-model-retrievers} can be re-written as:
\begin{align}
    \pth(\rmD | \rmQ) =  \frac{\ind[ \rmD \in \bm{\gT}_\phi^{(M)}] \exp F_\theta(\rmD, \rmQ)}{\sum_{\rmD' \in \bm{\gT}_\phi^{(M)}} \exp F_\phi(\rmD', \rmQ)}
    \quad
    \rphi(\rmD | \rmQ) =  \frac{\ind[ \rmD \in \bm{\gT_\phi}^{(M)}] \exp F_\phi(\rmD, \rmQ)}{\sum_{\rmD' \in \bm{\gT}_\phi^{(M)}} \exp F_\phi(\rmD', \rmQ)} 
    \ .
\end{align}

where $\bm{\gT}_\phi^{(M)} \defeq \gT_\phi(\rvq_1) \times \ldots \times \gT_\phi(\rvq_M)$ the set of combinations of $M$-document vectors ($P^{M}$ combinations).

\paragraph{VOD} By applying the results from section \ref{apdx:vod-objective} to $\rva_\star, \rmD, \rmQ$ with $\zeta(\rmD) = {\exp F_\theta(\rmD, \rmQ)} / {\exp F_\phi(\rmD, \rmQ)}$ the VOD objective and its gradient are:
\begin{subequations}
\begin{gather}
\renewcommand{\arraystretch}{1.5} 
\hat{L}_\alpha^{K}(\rva_\star, \rmQ) \defeq 
    \frac{1}{1-\alpha} \log \sum_{\rmD \in \sS^{(M)}} s(\rmD) \left( \frac{\zeta(\rmD) \pth(\rva_\star| \rmD, \rmQ)}{\sum_{\rmD' \in \sS^{(M)}} s(\rmD') \zeta(\rmD')} \right)^{1-\alpha}\\
     \mu_{\theta, \alpha, K}^\mathrm{VOD}(\rva_\star, \rmQ) \defeq 
     \sum_{\rmD \in \sS^{(M)}} \frac{s(\rmD) \left( \zeta(\rmD) \pth(\rva_\star | \rmD, \rmQ) \right)^{1-\alpha}}{\sum_{\rmD' \in \sS^{(M)}} s(\rmD') \left( \zeta(\rmD') \pth(\rva_\star | \rmD', \rmQ) \right)^{1-\alpha}} \left( \nabla_\theta\log \pth(\rmA | \rmD, \rmQ) + \rvh(\rmD' | \rmQ) \right) \ .
\end{gather}
\end{subequations}

where we define (we discuss the product of priority sampling estimates in Appendix \ref{apdx:priority-sampling})
\begin{subequations}
    \begin{gather}
        s(\rmD) \defeq \prod_{j=1}^M s_j[\rmD_j]) \\
        (\rvd_{j,1} , s_j[\rvd_1]), \ldots, (\rvd_{j,K}, s_j[\rvd_K]) \prioritysim \rphi(\rvd | \rvq_j) \,, \\
        \sS_j \defeq \{\rvd_{j,1}, \ldots, \rvd_{j,K} \}  \,, \quad
        \sS^{(M)} \defeq \sS_1 \times \ldots \times \sS_M \,.
    \end{gather}
\end{subequations}

\paragraph{Monte-Carlo estimation}\label{sec:mc-estimation}

During training, the computational budget is tight, and the VOD objective and its gradient are evaluated using a single set of samples $\sS^{(M)}$. During evaluation, we can leverage $C \geq 1$ Monte-Carlo samples $\sS^{M}_1, \ldots, \sS^{M}_C$, each containing $K^M$ document combinations sampled from $\rphi(\rmD | \rmQ)$ without replacement, to estimate the RVB (and therefore the log-likelihood) more accurately. We use the following estimate:
\begin{subequations}
\begin{gather}\label{eq:mc-estimate}
    \hat{p}_\theta(\rva, \rmQ) \defeq \frac{1}{C} \sum_{i=1}^C \frac{
    \exp \hat{L}_\alpha^{K}(\rva, \rmQ | \sS^{(M)}_i)
    }{\sum_{\rva' \in \rmA} \exp 
    \hat{L}_\alpha^{K}(\rva', \rmQ | \sS^{(M)}_i)
    }
    \\
    \sS^{(M)}_i \prioritysim \rphi(\rmD | \rmQ)\,, \quad \text{for } i \in [1, C] \,.
\end{gather}
\end{subequations}

\section{Implementation}\label{apdx:implementation}
\begin{table}[h]
  \caption{
  Parameterization of the reader and retriever scores. The complexity is reported for a batch-size of one, $M$ answer option, and for $K$ documents and inputs $\rvq_j = [\rvq; \rva_j]$ and $\rvd$ of lengths $L_\rvq$ and $L_\rva$. When using a dual-encoder architecture, the parameters of he BERT backbone are shared across the two encoders.}
  \label{tab:parameterization}
  \vskip 0.15in
  \begin{center}
  \begin{small}
  \renewcommand{\arraystretch}{1.5} 
  \begin{tabular}{lll}
\toprule
\bf Type & \bf Complexity & \bf Parameterization \\
\midrule
dual-encoder & $M(L^2_\rvq + K L^2_\rvd)$ & $ \fth(\rvd, \rvq_j) = \linear_{\theta[D]}(\bert_\theta(\rvd))^T \linear_{\theta[Q]}(\bert_\theta(\rvq_j))$ \\
Cross attn. & $MK (L_\rvq + L_\rvd)^2$ & $ \gth(\rvd, \rvq_j) = \linear_\theta(\bert_\theta([\rvd; \rvq_j]))$ \\

\bottomrule
\end{tabular}{}
  \end{small}
  \end{center}
  \vskip -0.10in
\end{table}

\paragraph{Documents preprocessing}

We encode the text and title of all the articles using the relevant BERT tokenizer. For each article with encoded title $\rvt$ of length $L_\rvt$, we extract overlapping passages $\rvp$ of length $L_\rvp = 200 - 2 - L_\rvt$ with stride 100 tokens. For each passage, using $\texttt{[DOC]}$ a special token added to the BERT vocabulary, we format each passage as 
\begin{equation}
    \rvd \defeq [\text{\texttt{[CLS]}}\ ;\ \text{\texttt{[DOC]}}\ ;\ \rvt\ ;\ \rvp] \ .
\end{equation}

\paragraph{Queries preprocessing}

We encode all questions and answer options using the tokenizer and store the question-answer pairs as 
\begin{equation}
\rvq_j \defeq [\text{\texttt{[CLS]}}\ ;\ \text{\texttt{[QUERY]}}\ ;\ \rvq\ ;\ \text{\texttt{[SEP]}}\ ;\   \rva_j]
\end{equation}
where the question $\rvq$ is truncated such as $|\rvq_j| \leq 312$ tokens and \texttt{[QUERY]} is an additional special token. On the reader side, we append the document passage $\rvd$ to the question-answer query $q_j$ such that $
\rvq_j \defeq [\rvd; \text{\texttt{[SEP]}}; \text{\texttt{[QUERY]}}; \rvq;\text{\texttt{[SEP]}}; \rva_j]$.

\paragraph{Reader}

We parameterize the reader score $\gth$ using a cross-attention model parameterized by another BERT backbone. Each query $\rvq_j=[\rvq; \rva_j]$ is prepended with a document $\rvd$, and an additional linear layer is used to reduce the output of BERT at the \texttt{CLS} token to a scalar value, as originally done in \cite{Devlin2018-qr}. See expression in Table \ref{tab:parameterization}.


\paragraph{Retriever}

We parameterize the retriever score $\fth$ using a dual encoder architecture similar to DPR, except that we share the BERT backbone across the two columns and one linear layer to project the output of each column. See expression in Table \ref{tab:parameterization}.

\paragraph{Hyperparameters}\label{sec:hyperparameters}
We summarize the training, evaluation and model hyperparameters in Table \ref{tab:hyperparameters}.

\section{Additional experimental data}\label{apdx:experimental-results}
\begin{figure}[H]
\begin{center}
\centerline{\includegraphics[width=\columnwidth]{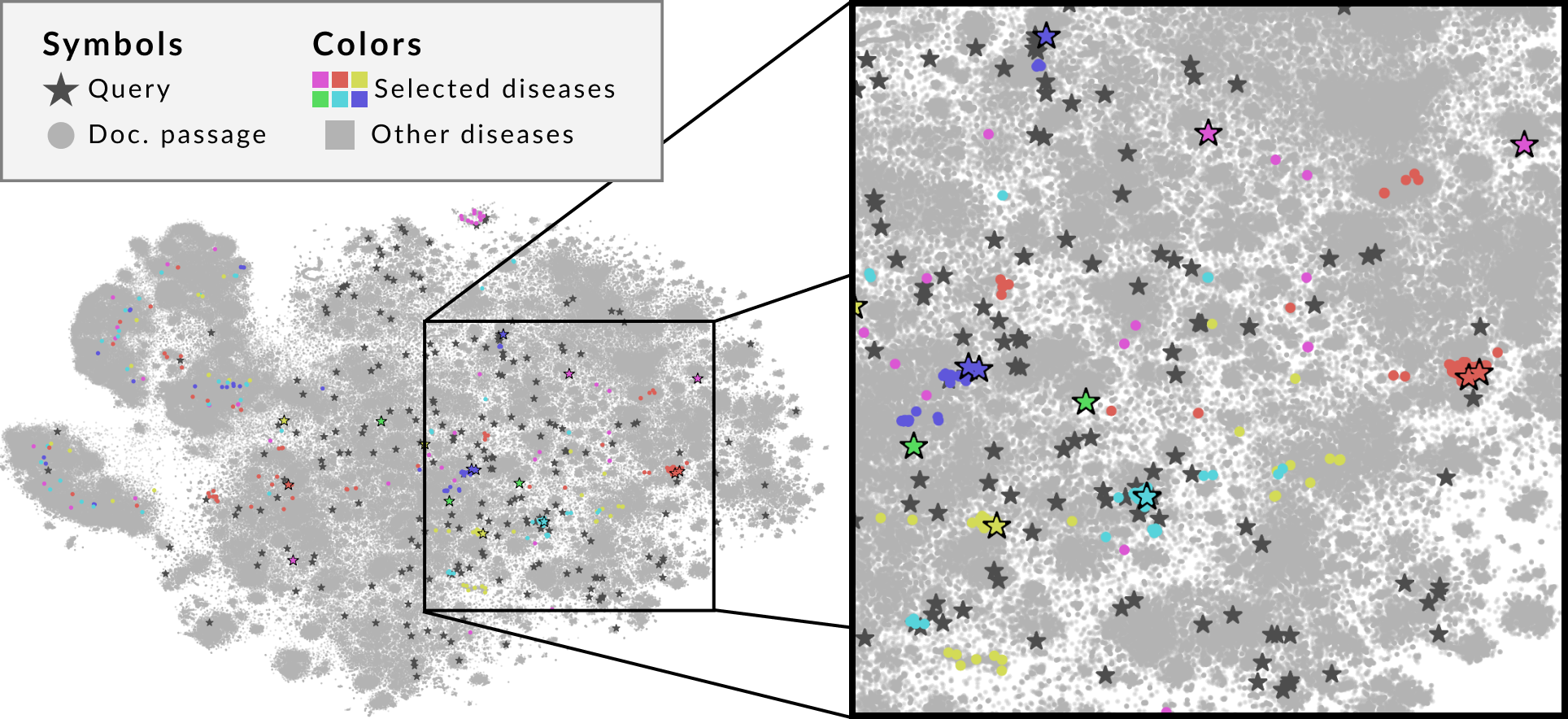}}
\caption{Visualizing the latent retrieval space. T-SNE projection of the embedding space where are encoded the 712k document passages of the FindZebra corpus and the 248 FindZebra queries. The documents and questions are annotated based on their disease identifier. The documents and queries annotated with the top 6 most frequent diseases (found in the queries) are highlighted with colours. The others are represented in gray. Some queries are successfully matched with a neighbourhood of relevant passages, although passages taken from a single document might be scattered across the embedding space.}
\label{fig:t-sne}
\end{center}
\vskip -0.2in
\end{figure}

\begin{SCfigure}[50][h]
  \centering
  \includegraphics[width=0.3\textwidth]{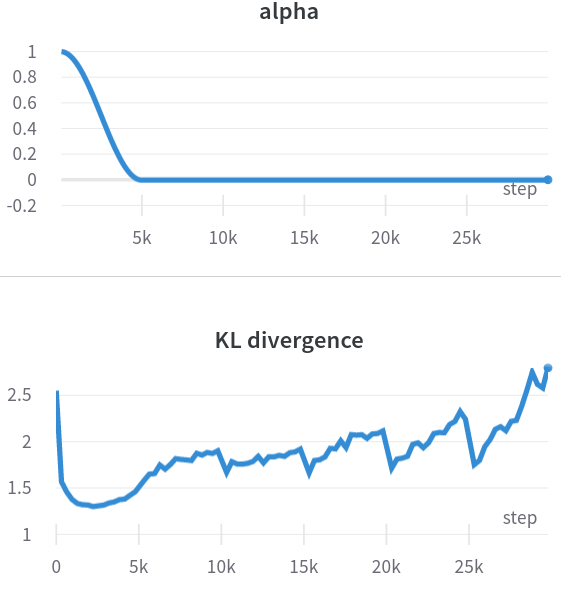}
  \caption{Measure of the divergence $\KL\left(\rphi(\rvd | \rvq)\ ||\ \pth(\rvd | \rvq) \right)$ during the training of a VOD retriever on the USMLE dataset. The retriever checkpoint is updated every $T=5k$ steps. $\alpha$ is annealed from 1 to 0 during the first 5k steps. We recognize the pattern schematized in Figure \ref{fig:periodic-training}. In this example, the approximate posterior is chosen as a combination of a checkpoint of the retriever and a static BM25 component. Therefore the value of the divergence is never zero because the divergence between the model and the BM25 retriever is always strictly positive.}
  \label{fig:apdx-experimental-kl}
\end{SCfigure}

In Table \ref{tab:retrieval-samples}\ref{tab:retrieval-samples}, we report retrieved top-1 passages for the distilled retriever (two successes and two failures). In Figure \ref{fig:apdx-experimental-kl}, we report the measurement of the $\KL\left(\rphi(\rvd | \rvq)\ ||\ \pth(\rvd | \rvq) \right)$ during training of a VOD model. In Figure \ref{fig:t-sne}, we illustrated the FindZebra queries and corpus embedded using the trained BioLinkBERT model and projected using t-SNE.

\subsection{Retrieval samples}\label{apdx:retrieval-samples}

\begin{table}[H]
  \caption{Top-1 passages retrieved for a selection of FindZebra queries with their annotated answer CUIs and the rank of the first matching article for VOD (BioLinkBERT onnly, with distillation) and the FindZebra API. We showcase the retriever model trained with task-specific distillation and without BM25 coupling (MRR 31.7). We highlight terms from the queries and passages relevant to each other.}
  \label{tab:retrieval-samples}
  \centering
  \vskip 0.15in
  \resizebox{\columnwidth}{!}{%
  \begin{tabular}{ccc}
    \toprule
     & \bf Query & \bf Top-1 passage (VOD, BioLinkBERT backbone, with distillation) \\
    \midrule
     1 & \multicolumn{1}{p{0.4\linewidth}}{
    Q: \textbf{widespread musculoskeletal pain for more than 6 months and point tenderness in at least 11 of 18 defined anatomical sites}
    \newline
    A: Fibromyalgia (C0016053)
    \newline
    \textit{Hit rank: $\mathrm{VOD}_\mathrm{BioLinkBERT}$=1, $\mathrm{FZ}_\mathrm{API}$=1}
    }
    &
    \multicolumn{1}{p{1.0\linewidth}}{
    \textit{Fibromyalgia.} (...) for IL-1 receptor antagonist, IL-6 and IL-8.
    
Diagnosis
  The location of the nine paired tender points that comprise the 1990 American College of Rheumatology criteria for fibromyalgia
There is no single pathological feature, laboratory finding or biomarker that can diagnose fibromyalgia and there is debate over what should be considered diagnostic criteria and whether an objective diagnosis is possible. In most cases, people with fibromyalgia symptoms may have laboratory test results that appear normal and many of their symptoms may mimic those of other rheumatic conditions such as arthritis or osteoporosis. The most widely accepted set of classification criteria for research purposes was elaborated in 1990 by the Multicenter Criteria Committee of the American College of Rheumatology. These criteria, which are known informally as "the ACR 1990", define fibromyalgia according to the presence of the following criteria:
\textbf{A history of widespread pain lasting more than three months – affecting all four quadrants of the body, i.e., both sides, and above and below the waist}.
Tender points – there (...)
} 
\\
\midrule
2 &
\multicolumn{1}{p{0.4\linewidth}}{
Q: diagnosis for dementing syndrome characterized primarily by \textbf{impairment of interpersonal and executive function}
\newline
A: Frontotemporal dementia (C0338451)
\newline
\textit{Hit rank: $\mathrm{VOD}_\mathrm{BioLinkBERT}$=1, $\mathrm{FZ}_\mathrm{API}$=8}

}
&
\multicolumn{1}{p{1.0\linewidth}}{
\textit{Frontotemporal dementia.} (FTDs) are a group of neurodegenerative disorders associated with shrinking of the frontal and temporal anterior lobes of the brain. \textbf{Symptoms include marked changes in social behavior and personality, and/or problems with language}. People with behavior changes may have disinhibition (with socially inappropriate behavior), apathy and loss of empathy, hyperorality (eating excessive amounts of food or attempting to consume inedible things), agitation, compulsive behavior, and various other changes. Examples of problems with language include difficulty speaking or understanding speech. \textbf{Some people with FTD also develop a motor syndrome} such as parkinsonism or motor neuron disease (which may be associated with various additional symptoms).

There is a strong genetic component to  FTDs. It sometimes follows an autosomal dominant inheritance pattern, or sometimes there is a general family history of dementia or psychiatric disorders. The three main genes responsible for familial FTD are MAPT, GRN, and C9orf72. However, the (...)
}
\\
\midrule
3 &
\multicolumn{1}{p{0.4\linewidth}}{
Q: syndrome characterized by \textbf{cough, reversible wheezing, and peripheral blood eosinophilia}
\newline
A: Asthma (C0004096), Reactive airway disease (C3714497)
\newline
\textit{Hit rank: $\mathrm{VOD}_\mathrm{BioLinkBERT}$=72, $\mathrm{FZ}_\mathrm{API}$=11}
}
&
\multicolumn{1}{p{1.0\linewidth}}{
\textit{L\"{o}ffler's syndrome.} (...)
a parasitic infection such as irritable bowel syndrome, abdominal pain and cramping, skin rashes and fatigue. L\"{o}ffler's syndrome itself will cause difficulty breathing, coughing as well as a fever.

Contents 
1 Diagnosis
2 Prevention
3 Epidemiology
4 History
5 See also
6 References
7 External links

Diagnosis
The diagnosis of Loffler's syndrome can be challenging, as the diagnostic criteria can be vague and consistent with a multitude of diseases or conditions. The disease's developmental trajectory is mostly unknown. Upon examination of symptoms, a doctor will likely request a chest x-ray looking for migratory pulmonary infiltrate, and blood testing, to confirm a diagnosis. Symptoms tend to be brief, but can range from mild to severe and include: fever, vomiting, increased \textbf{respirations or difficulty breathing, cough, wheeze}, and rash. Symptoms typically follow an exposure to allergens or certain drugs, and last approximately two weeks.
\textbf{Eosinophilia is the main feature of diagnostic} (...)
}
\\
\midrule
4 &
\multicolumn{1}{p{0.4\linewidth}}{
Q: 5 year old, boy, congenital malformations, \textbf{malformations of the hands} and feet, bilateral strabismus, small tongue, impaired coordination, expressionless face, \textbf{prominent forehead}, depressed nasal bridge, hypoplastic thumbs, bilateral adactyly of the feet, \textbf{short stature}, severe myopia  
\newline
{A: Mobius Syndrome (C0221060), Mobius II syndrome (C0853240)}
\newline
\textit{Hit rank: $\mathrm{VOD}_\mathrm{BioLinkBERT}$=$\infty$, $\mathrm{FZ}_\mathrm{API}$=1}
}
&
\multicolumn{1}{p{1.0\linewidth}}{
\textit{Achondroplasia.} 
(...) hypochondroplasia, but the features of achondroplasia tend to be more severe. \textbf{All people with achondroplasia have short stature}. The average height of an adult male with achondroplasia is 131 centimeters (4 feet, 4 inches), and the average height for adult females is 124 centimeters (4 feet, 1 inch). Characteristic features of achondroplasia include an average-size trunk, short arms and legs with particularly short upper arms and thighs, limited range of motion at the elbows, and an enlarged head (macrocephaly) with a \textbf{prominent forehead}. \textbf{Fingers are typically short and the ring finger and middle finger may diverge, giving the hand a three-pronged (trident) appearance}. People with achondroplasia are generally of normal intelligence. Health problems commonly associated with achondroplasia include episodes in which breathing slows or stops for short periods (apnea), obesity, (...)
}\\
\bottomrule
\end{tabular}
  }
\end{table}

\subsection{Embedding space}\label{apdx:t-SNE}

\subsection{Empirical divergence measured during training}\label{apdx:empirical-kl}

\sidecaptionvpos{figure}{c}

\clearpage
\newpage
\section{MedWiki}\label{apdx:medwiki}

\begin{table}[!h]
  \caption{Comparing the MedWiki with the original MedQA corpus on the USMLE dataset.}
  \label{tab:ablation_corpus_results}
  \vskip 0.15in
  \centering
  \begin{tabular}{llll|rr}
    \toprule
    \bf Method & \bf Reader & \bf Retriever & \bf Corpus & \bf Valid. & \bf Test \\
    \midrule
    Disjoint & BioBERT\textsuperscript{1} & BM25 & MedQA\textsuperscript{2} & 37.68 & 39.54 \\
    Disjoint & BioBERT\textsuperscript{1} & BM25  & MedWiki  & 38.82 & 40.46 \\
    Disjoint & BioLinkBERT & BM25  & MedQA\textsuperscript{2}  &  40.37 & 41.05 \\
    Disjoint & BioLinkBERT & BM25  & MedWiki  &  \bf 42.21 & \bf 42.25 \\
    \bottomrule
     \addlinespace[0.1cm]
    \multicolumn{6}{l}{\small
    \textsuperscript{1}model weights from \cite{Lee2020-qq},
    ~~\textsuperscript{2}original corpus from \cite{Jin2021-jo}
    }
  \end{tabular}
  \vskip 0.1in
\end{table}

The MedWiki corpus is a set of Wikipedia articles collected for research on medical question answering with low resources. Existing medical corpora, such as the MedQA corpus, are not adequately aligned with the ODQA task and are often measly and fragmented. At the same time, all of Wikipedia is cumbersome to use on consumer hardware. In order to reflect the true information need of medical experts, we assembled the MedWiki corpus by using real-world medical entrance exam questions. We queried the Wikipedia API using the answer options from all dataset splits of USMLE and MedMCQA and retained the top-10 articles for each answer option. This corpus includes 293.6k unique Wikipedia articles ($\approx4.5$\% of Wikipedia) that cover a broad range of medical topics.



\subsection*{MedQA vs. MedWiki}

In the following paragraph, we compare the MedWiki corpus with the original MedQA corpus \cite{Jin2021-jo}.



\paragraph{Qualitative comparison} 

Using ElasticSearch, we compare the retrieved documents of MedWiki to the ones of MedQA. In Table \ref{tab:medqa_vs_wiki}, \ref{tab:medqa_vs_wiki_2}, \ref{tab:medqa_vs_wiki_3} we present a few examples. The MedQA corpus is a selection of medical textbooks which often revolve around medical case studies, akin to the USMLE questions (see example in Table \ref{tab:medqa_vs_wiki}). In contrast, the MedWiki corpus references Wikipedia articles which are often edited to be concise, which is especially true for the abstract part of the articles, which contain the basic and usually most important information about a topic. Furthermore, each Wikipedia article comes with a title, which augments each passage with a higher-level context.

However, our approach of querying against the Wikipedia API results in many out-of-domain articles. For instance in Table \ref{tab:medqa_vs_wiki_2}, we display a MedWiki passage that originates from a non-medical article. Although the MedQA corpus is strictly oriented toward medical topics, it was built by extracting text from physical books using OCR software, which led to errors in the process and ultimately resulted in part of the corpus being unreadable.

Overall, both corpora provide adequate evidence to answer USMLE questions. Nevertheless, the MedWiki corpus is three times larger in vocabulary size and eight times more extensive in word count, making it more robust and diverse.


\paragraph{Quantitative comparison} 

We investigated how the two corpora affect the final QA accuracy on the USMLE dataset. In contrast with the rest of the paper, we used a multi-document reader, as done in \cite{Jin2021-jo}. We used an ElasticSearch index to retrieve the set of top 3 documents $\{\rvd_1, \rvd_2, \rvd_3 \}$ for each pair ($\rvq, \rva_i$) as context for each answer option. The normalized log probabilities over the four options were obtained by processing the set of concatenated tokens $[\rvd_1;\rvd_2;\rvd_3;\rvq; \rva_i]$ with BERT. We performed all experiments using a batch size of 16, set the learning rate to 1e-5, and run all experiments for 30 epochs. We report the predictive accuracy averaged for three initial random seeds.

Table \ref{tab:ablation_corpus_results} summarizes the performance on the two corpora. We see that our collected MedWiki corpus leads to better QA performance by 0.9\%-1.2\% absolute. This result indicates that the MedWiki corpus can safely be used as a replacement of the MedQA corpus. The MedWiki yields USMLE accuracy that is superior to using the MedQA corpus (Table \ref{tab:ablation_corpus_results}), and yields good results on the MedMCQA (Table \ref{tab:odqa-medmcqa-accuracy}) despite consisting in only of a fraction of the English Wikipedia.

\renewcommand{\arraystretch}{1.5}
\begin{table*}[ht]
\tiny
\centering
\captionsetup{font=scriptsize}
\begin{tabular}{p{1.2cm}|p{12cm}}
        \hline
        \textbf{Question} & a 5 year old girl is brought to the emergency department by her mother because of multiple episodes of nausea and vomiting that last about 2 hours. during this period she has had 6 8 episodes of bilious vomiting and abdominal pain. the vomiting was preceded by fatigue. the girl feels well between these episodes. she has missed several days of school and has been hospitalized 2 times during the past 6 months for dehydration due to similar episodes of vomiting and nausea. the patient has lived with her mother since her parents divorced 8 months ago. her immunizations are up to date. she is at the 60th percentile for height and 30th percentile for weight. she appears emaciated. her temperature is 36. 8 c 98. 8 f pulse is 99 min and blood pressure is 82 52 mm hg. examination shows dry mucous membranes. the lungs are clear to auscultation. abdominal examination shows a soft abdomen with mild diffuse tenderness with no guarding or rebound. the remainder of the physical examination shows no abnormalities. which of the following is the most likely diagnosis?
 \tabularnewline
        \hline
        \textbf{Options} & \textbf{A: cyclic vomiting syndrome}, B: gastroenteritis, C: hypertrophic pyloric stenosis, D: gastroesophageal reflux disease
        \tabularnewline
        \hline
        \hline
        \textbf{Document} \par \textbf{from} \par \textbf{MedQA} & headache, and sweating patient presentation : be is a 45 - year - old woman who presents with concerns about sudden ( paroxysmal ), intense, brief episodes of headache, sweating (diaphoresis), and a racing heart (palpitations). focused history : be reports that the attacks started ~ 3 weeks ago. they last from 2 to 10 minutes, during which time she feels quite anxious. during the attacks, it feels as though her heart is skipping beats (arrhythmia). at first, she thought the attacks were related to recent stress at work and maybe even menopause. the last time it happened, she was in a pharmacy and had her blood pressure taken. she was told it was 165 / 110 mm hg. be notes that she has lost weight ($\sim$8 lbs) in this period even though her appetite has been good. pertinent findings : the physical examination was remarkable for be ’ s thin, pale
        \tabularnewline
        \hline
        \textbf{Document} \par \textbf{from} \par \textbf{MedWiki} & \textbf{panayiotopoulos syndrome}. pital, or calcarine sulci. follow - up meg demonstrated shifting localization or disappearance of meg spikes. illustrative cases in a typical presentation of panayiotopoulos syndrome, the child looks pale, vomits, and is fully conscious, able to speak, and understand but complains of “ feeling sick. ” two thirds of the seizures start in sleep ; the child may wake up with similar complaints while still conscious or else may be found vomiting, conscious, confused, or unresponsive. case 1. a girl had 2 seizures in sleep at 6 years of age. in the first fit she was found vomiting vigorously, eyes turned to one side, pale, and unresponsive. her condition remained unchanged for 3 hours before she developed generalized tonic - clonic convulsions. she gradually improved, and by the next morning was normal. the second seizure occurred 4 months later. she awoke and told her mother that she wanted to vomit, 
        \tabularnewline
        \hline
        \hline
    \end{tabular}
\caption{An example of the retrieved documents from the MedQA and MedWiki corpus respectively. Correct answers and document titles are highlighted when available.}
\label{tab:medqa_vs_wiki}
\end{table*}

\begin{table*}[h]
\tiny
\centering
\captionsetup{font=scriptsize}
\begin{tabular}{p{1.2cm}|p{12cm}}
        \hline
        \textbf{Question} & a 40 year old woman presents with difficulty falling asleep 
diminished appetite and tiredness for the past 6 weeks. she says that despite going to bed early at 
night she is unable to fall asleep. she denies feeling anxious or having disturbing thoughts while in 
bed. even when she manages to fall asleep she wakes up early in the morning and is unable to fall back 
asleep. she says she has grown increasingly irritable and feels increasingly hopeless and her 
concentration and interest at work have diminished. the patient denies thoughts of suicide or death. 
because of her diminished appetite she has lost 4 kg 8. 8 lb in the last few weeks and has started 
drinking a glass of wine every night instead of eating dinner. she has no significant past medical 
history and is not on any medications. which of the following is the best course of treatment in this 
patient?
 \tabularnewline
        \hline
        \textbf{Options} & A: diazepam, B: paroxetine, C: zolpidem, \textbf{D: trazodone}
        \tabularnewline
        \hline
        \hline
        \textbf{Document} \par \textbf{from} \par \textbf{MedQA} &
headache, and sweating patient presentation : be is a 45 - year - old woman who presents with concerns about sudden ( paroxysmal ), intense, brief episodes of headache, sweating (diaphoresis), and a racing heart (palpitations). focused history : be reports that the attacks started ~ 3 weeks ago. they last from 2 to 10 minutes, during which time she feels quite anxious. during the attacks, it feels as though her heart is skipping 
beats (arrhythmia). at first, she thought the attacks were related to recent stress at work and maybe even menopause. the last time it happened, she was in a pharmacy and had her blood pressure taken. she was told it was 165 / 110 mm hg. be notes that she has lost weight ($\sim$8 lbs) in this period even though her appetite has been good. pertinent findings : the physical examination was remarkable for be ’ s thin, pale 
        \tabularnewline
        \hline
        \textbf{Document} \par \textbf{from} \par \textbf{MedWiki} & \textbf{hillary clinton's tenure as secretary of state}. hillary to the middle east to talk about how these countries can transition to new leaders — though, i've got to be honest, she's gotten a little passionate about the subject. these past few weeks it's been tough falling asleep with hillary out there on pennsylvania avenue shouting, throwing rocks at the window. in any case, obama's reference to clinton travelling a lot was true enough ; by now she had logged in her boeing 757, more than any other secretary of state for a comparable period of time, and had visited 79 countries while in the office. time magazine wrote that "clinton's endurance is legendary" and that she would still be going at the end of long work days even as her staff members were glazing out. the key was her ability to fall asleep on demand, at any time and place, for power naps. clinton also saw the potential political changes in the mideast as an opportunity for an even more fundamental change
        \tabularnewline
        \hline
        \hline
    \end{tabular}
\caption{An example of the two different retrieved documents from the MedQA and MedWiki corpus. Correct answers and document titles are highlighted when available.}
\label{tab:medqa_vs_wiki_2}
\end{table*}

\begin{table*}[!htbp]
\tiny
\centering
\captionsetup{font=scriptsize}
\begin{tabular}{p{1.2cm}|p{12cm}}
        \hline
        \textbf{Question} & a 37 year old female with a history of type ii diabetes mellitus presents to the emergency department complaining of blood in her urine left sided flank pain nausea and fever. she also states that she has pain with urination. vital signs include temperature is 102 deg f 39. 4 deg c blood pressure is 114 82 mmhg pulse is 96 min respirations are 18 and oxygen saturation of 97 on room air. on physical examination the patient appears uncomfortable and has tenderness on the left flank and left costovertebral angle. which of the following is the next best step in management?
 \tabularnewline
        \hline
        \textbf{Options} & A: obtain an abdominal ct scan, \textbf{B: obtain a urine analysis and urine culture}, C: begin intravenous treatment with ceftazidime, D: no treatment is necessary
        \tabularnewline
        \hline
        \hline
        \textbf{Document} \par \textbf{from} \par \textbf{MedQA} & rim, \& quinolones camille e. beauduy, pharmd, \& lisa g. winston, md $^{\star}$ a 59 - year - old woman presents to an urgent care clinic with a 4 - day history of frequent and painful urination. she has had fevers, chills, and flank pain for the past 2 days. her physician advised her to come immediately to the clinic for evaluation. in the clinic she is febrile (38. $5^{\circ}$c [ 101. $3^{\circ}$f ]) but otherwise stable and states she is not experiencing any nausea or vomiting. her urine dipstick test is positive for leukocyte esterase. urinalysis and urine culture are ordered. her past medical history is significant for three urinary tract infections in the past year. each episode was uncom - plicated, treated with trimethoprim - sulfamethoxazole, and promptly resolved. she also has osteoporosis
        \tabularnewline
        \hline
        \textbf{Document} \par \textbf{from} \par \textbf{MedWiki} & \textbf{hydronephrosis}. hydronephrosis describes dilation of the renal pelvis and calyces as a result of obstruction to urine flow. signs and symptoms the signs and symptoms of hydronephrosis depend upon whether the obstruction is acute or chronic, partial or complete, unilateral or bilateral. hydronephrosis that occurs acutely with sudden onset (as caused by a kidney stone) can cause intense pain in the flank area (between the hips and ribs). historically, this type of pain has been described as "dietl's crisis". conversely, hydronephrosis that develops gradually will generally cause either a dull discomfort or no pain. nausea and vomiting may also occur. an obstruction that occurs at the urethra or bladder outlet can cause pain and pressure resulting from distension of the bladder. blocking the flow of urine will commonly result in urinary tract infections which can lead to the development of  stones, fever, and blood or pus in the urine
        \tabularnewline
        \hline
        \hline
    \end{tabular}
\vspace{1mm}
\caption{An example of the two different retrieved documents from the MedQA and MedWiki corpus. Correct answers and document titles are highlighted when available.}
\label{tab:medqa_vs_wiki_3}
\end{table*}

\clearpage
\newpage

\begin{table}[!h]
  \caption{Hyperparameters used across the multiple-choice ODQA experiments.}
  \label{tab:hyperparameters}
  \vskip 0.15in
  \begin{center}
  \begin{scriptsize}
    \begin{tabular}{lll}
\toprule
\bf Category & \bf Parameter & \bf Value \\
\midrule
Optimization & Optimizer & AdamW \\
&Learning rate & $3 \cdot 10^{-6}$ \\
&Learning rate warmup & $0.1 \cdot T$\\
&Warmup frequency & every $T$ steps \\
&Weight decay & $1 \cdot 10^{-3}$ \\ 
&Gradient clipping & 0.5\\
&Precision & \texttt{float16}  \\
\midrule
$\alpha$ annealing & initial value & 1 \\
& final value & 0 \\
& length & $T$ steps \\
& type & cosine \\
\midrule
Model&Reader & BioLinkBERT + linear layer \\
&Retriever & BioLinkBERT + two linear layers \\
&Output vector size & 768 \\
\midrule
Batching&batch-size & 32 \\
&$M$ (\# of options) & 4 \\
&$K$ (documents per option) & 8 \\
&$P$ (retriever support size) & 100 \\
&$N$ (corpus size) & 7,766.9k \\
&document passage stride & 100 \\
&$L_\rvd$ (document passage length) & 200 \\
&max. $L_\rvq$ (max. query length) & 312 \\
&max. $L_\rvd + L_\rvq$  & 512 \\
\midrule
Training &$T$ (re-indexing period length) & 5k \\
&Training steps (MedMCQA) & 150k \\ 
&Training steps (USMLE) & 50k \\ 
&Training steps (MedMCQA $\rightarrow$USMLE) & 150k $\rightarrow$ 10k \\
&Training steps (Distillation) & 120k \\
\midrule
Posterior and retrieval& parameterization & $\fphi^{\rm{ckpt}}(\rvd, [\rvq; \rva]) + \tau^{-1} \left( \rm{BM25}(\rvq) + \beta \cdot \rm{BM25}(\rva) \right) $ \\
&$\tau$ (BM25 temperature) & 5 \\
&$\beta$ (BM25 answer weight) & $1 + 0.5\ \rm{max}\left\{0, \log \left( \nicefrac{L_\rvq}{L_\rva}\right) \right\}$ \\
& BM25 implementation & \texttt{elasticsearch} v7.14.1 \\
& BM25 paramters & b=0.75, k1=1.2 \\
& MIPS implementation & \texttt{faiss v1.7.2} \\
&\texttt{faiss} factory string & IVF1000,Flat \\
&\texttt{faiss} precision & \texttt{float16} \\
&\texttt{faiss} nprobe & 32 \\
\midrule
Evaluation&$C$ (Monte-Carlo samples for eval.) & 10 \\
\midrule
Hardware& CPU & AMD EPYC 7252 8-Core Processor \\
& RAM & 256 GB \\
& GPU & 8 $\times$ Quadro RTX 5000 \\
& VRAM & 128 GB \\
\midrule
Software & PyTorch & \cite{Paszke2019-lk}\\
& Lightning & \cite{Falcon_undated-fa} \\
& \texttt{faiss} & \cite{Johnson2021-do}\\
\bottomrule
\end{tabular}
  \end{scriptsize}
  \end{center}
  \vskip -0.1in
\end{table}

\begin{table}[!h]
  \caption{Mathematical symbols.}
  \label{tab:symbols}
  \vskip 0.15in
  \begin{center}
  \resizebox*{!}{0.95\textheight}{
  \begin{tiny}
    \begin{tabular}{lll}
\toprule
\bf Category & \bf Symbol & \bf Description \\
\midrule
ODQA variables & $\rva$ & answer\\
&$\rvd$ & document or document passage\\
&$\rvq$ & question or query\\
& $L_\rva$ & number of tokens in the answer \\
& $L_\rvd$ & number of tokens in the document \\
& $L_\rvq$ & number of tokens in the query \\
&$\sD$ & corpus of documents \\
&$N$ & number of documents in the corpus\\
\midrule
Reader-retriever & $\theta$ & parameter of the retrieval-augmented model (generative model)\\
& $\pth(\rva, \rvd | \rvq)$ & Joint reader-retriever model\\
& $\wall(\rva, \rvq, \rvd)$ & Importance weight\\
& $\hat{v}_{\theta, \phi}(\rva, \rvq, \rvd)$ & Self-normalized importance weight estimate\\
& $\zeta(\rvd)$ & un-normalized density ratio $\propto \pth(\rvd | \rvq) \rphi^{-1}(\rvd |\rva, \rvq)$ \\ 
& $\pth(\rva| \rvd, \rvq)$ & reader \\
& $\pth(\rvd |\rvq)$ & retriever \\
& $\fth(\rvd, \rvq)$ & score of the retriever \\
\midrule
Posterior & $\phi$ & parameter of the approximate posterior (inference network)\\
& $\rphi(\rvd |\rva, \rvq)$ & approximate posterior (static retriever) \\
& $\fphi(\rva, \rvd, \rvq)$ & score of the approximate posterior \\
& $\mathrm{BM25}(\rvq, \rvd)$ & BM25 score of the query $\rvq$ for the document $\rvd$ \\
& $\fphi^{\rm{ckpt}}(\rvd, \rvq)$ & checkpoint of the retriever \\
& $\tau$ & temperature balancing the checkpoint score and the BM25 score \\
& $\beta$ & weight balancing the query and answer options BM25 terms \\
\midrule
Truncated retriever & $P$ & number of documents with non-zero mass under $\pth(\rvd |\rvq)$ \\
& $\gT_\phi$ & set of top-$P$ documents ranked by $\fphi$ (retrievers support) \\
\midrule
Sampling & $ (\rvd_1, s_1), \ldots, (\rvd_K, s_K) \prioritysim p(\rvd)$ & priority sampling (without replacement) wth samples $\rvd_i$ and weights $s_i$ \\
& $s_1, \ldots, s_K $ & priority weights \\
& $K$ & number of document samples with $K \leq P \leq N$ \\
& $C$ & number of Monte-Carlo samples (evaluation) \\
\midrule
Bounds & $\log \pth(\rva, \rvq)$ & Marginal task likelihood \\ 
& $\VIB(\rva, \rvq)$ & Variational Lower bound (ELBO) \\ 
& $\RVB_\alpha(\rva, \rvq)$ & R\'enyi Variational Bound (RVB) \\
& $\RVB_\alpha^K(\rva, \rvq)$ & importance-weighted RVB (IW-RVB) \\
& $\alpha$ & parameter of the RVB \\
& $\hat{L}_{\alpha}^{K}(\rva, \rvq)$ & VOD objective (self-normalized importance sampling estimate of the RVB) \\
& $\mu_{\theta, \alpha, K}^{\mathrm{VOD}}$ & VOD gradient \\
& $\KL(\rphi(\rvd | \rva, \rvq)\|\pth(\rvd | \rva, \rvq))$ & KL divergence from the true posterior to the approximate posterior \\ 
& $\KL(\rphi(\rvd | \rva, \rvq)\|\pth(\rvd | \rvq))$ 
& KL divergence from the retriever to the approximate posterior \\ 
\midrule
Multiple-choice  & $\rva_i$ & answer option $i$ \\
& $\star$ & index of the correct answer option \\
& $\rvq_i$ & question-answer pair $[\rvq ; \rva_i]$\\
& $M$ & number of answer options \\
& $\rmA$ & vector of $M$ answer choices \\
& $\rmD$ & vector of $M$ documents \\
& $\rmQ$ & vector of $M$ queries (each expressed as $[\rvq; \rva_i]$) \\
& $\gth(\rvd, \rvq)$ & score of the reader (multiple-choice) \\
& $\sS^{(M)}$ & Cartesian product of the per-option samples $\sS_1, \ldots, \sS_M$ \\
& $\bm{\gT_\phi}^{(M)}$ & Product of the per-option top-$P$ sets $\gT_\phi(\rvq_1) \times, \ldots, \times \gT_\phi(\rvq_M)$ \\
\midrule
Spaces and Sets & $\Omega$ & space of strings \\
& $\sR$ & reals \\ 
& $(0,1]$ & real numbers in the interval [0, 1], 0 excluded \\
\midrule
Operators & $:=$ & defined as \\  & $[\cdot ; \cdot]$ & concatenation operator \\ 
& $\times$ & Cartesian product\\ 
& $\KL(p || q)$ & Kullback–Leibler (KL) divergence from $q$ to $p$\\
& $\ind[ \rvx \in \sX]$ & indicator function with value 1 if $\rvx \in \sX$ otherwise 0 \\ 
\bottomrule
\end{tabular}
  \end{tiny}
  }
  \end{center}
  \vskip -0.1in
\end{table}

\end{document}